\documentclass[manuscript]{acmart} 
\usepackage{algorithm}
\usepackage{algorithmic}
\usepackage{amsmath}
\usepackage{multirow} 
\usepackage{booktabs}
\usepackage{pdflscape}
\usepackage{longtable}
\usepackage{rotating}
\usepackage{colortbl}
\usepackage{boldline}
\AtBeginDocument{%
  \providecommand\BibTeX{{%
    \normalfont B\kern-0.5em{\scshape i\kern-0.25em b}\kern-0.8em\TeX}}}

\setcopyright{acmcopyright}
\copyrightyear{2018}
\acmYear{2018}
\acmDOI{XXXXXXX.XXXXXXX}

\acmPrice{15.00}
\acmISBN{978-1-4503-XXXX-X/18/06}

\begin{document}

\title{"Forgetting" in Machine Learning and Beyond: A Survey}

\author{Alyssa Sha}
\email{alyssa.sha@anu.edu.au}
\orcid{1234-5678-9012}
\affiliation{%
  \institution{Australian National University}
  \city{Canberra}
  \state{ACT}
  \country{Australia}
}

\author{Bernardo Pereira Nunes}
\email{Bernardo.Nunes@anu.edu.au}
\affiliation{%
  \institution{Australian National University}
  \city{Canberra}
  \state{ACT}
  \country{Australia}
}

\author{Armin Haller}
\email{Armin.Haller@anu.edu.au}
\affiliation{%
  \institution{Australian National University}
  \city{Canberra}
  \state{ACT}
  \country{Australia}
}

\begin{abstract}
This survey investigates the multifaceted nature of forgetting in machine learning, drawing insights from neuroscientific research that posits forgetting as an adaptive function rather than a defect, enhancing the learning process and preventing overfitting. This survey focuses on the benefits of forgetting and its applications across various machine learning sub-fields that can help improve model performance and enhance data privacy. Moreover, the paper discusses current challenges, future directions, and ethical considerations regarding the integration of forgetting mechanisms into machine learning models.
\end{abstract}




\maketitle

\section{Introduction}
Human brain is a complex system, where forgetting serves as a dynamic nature that allows us to avoid cognitive overload, update information to adapt to changing environments \cite{Gravitz2019}, and can potentially enhance our learning capabilities \cite{Bjork2019}. The advantages of forgetting have been investigated in various research fields, including education, philosophy, ecology and linguistics, where forgetting has been found to contribute significantly to the enhancement of humans' decision-making, creativity, and diversity from multiple perspectives.

Forgetting, an intrinsic aspect of human memory, does not naturally occur in machines, highlighting a fundamental distinction between humans and artificial systems. In the context of the human brain, overfitting arises when we simply memorise specific examples rather than generalise patterns from them \cite{Hoel2021}. This narrow focus can cause inflexibility in our thinking and problem-solving abilities, as well as lead to erroneous predictions or assumptions when confronted with unfamiliar situations. Overfitting is also a challenge in machine learning (ML) \cite{Dietterich1995}. By mimicking the human brain, incorporating a forget-and-relearn function into machines has been proposed to be a powerful paradigm for shaping the learning trajectories of artificial neural networks \cite{Zhou2022}, as not all content in the past is equally important for models to remember \cite{Sukhbaatar2021}.

There are different types of forgetting. Selective forgetting involves selectively ignoring irrelevant or noisy data. This form of forgetting aids in optimising the model's memory utilisation \cite{Sukhbaatar2021}, improving their generalisation ability \cite{Zhou2022}, and enhancing their adaptability to different datasets and tasks \cite{Zhang2023a}. Besides model performance, selective forgetting is reinforced by the principles of the General Data Protection Regulation (GDPR) in Europe and the California Consumer Privacy Act (CCPA) \cite{Bourtoule2021} to address the critical need for compliance with privacy laws and the ethical handling of personal data without the need for resource-intensive retraining \cite{Zhang2023}. On the other hand, detrimental forgetting occurs when a model loses previously learned information upon learning new information, a phenomenon known as catastrophic forgetting. This is particularly problematic in continuous learning environments, where a model needs to retain knowledge over time. Detrimental forgetting undermines the model's ability to build upon existing knowledge, including poor performance on previously learned tasks \cite{French1999}, the loss of information that could be useful for future tasks, and biased decision-making.

This survey article only focuses on selective forgetting, which acknowledges that forgetting certain information can be beneficial by allowing the model to prioritise and retain more important or relevant information, and protect user privacy. We then explore how theories and research findings on selective forgetting from other disciplines can be applied to the field of machine learning to enhance models' performance and address data privacy concerns. 

This paper surveys the forgetting in machine learning literature to answer the following research questions:

[RQ1]: How is forgetting manifested in different knowledge areas (e.g., psychology, philosophy, neuroscience)? This research question explores the multifaceted nature of forgetting to inspire the development of new forgetting models in machine learning approaches.

[RQ2]: How can forgetting be used to comply with data privacy laws, reduce bias and prioritise relevant information in Machine Learning approaches? 

[RQ3]: What are the future research opportunities and challenges associated with implementing forgetting mechanisms in machine learning? Here, we explore the current research gaps to advance this field.

\begin{figure}
    \centering
    \includegraphics[width=0.85\linewidth]{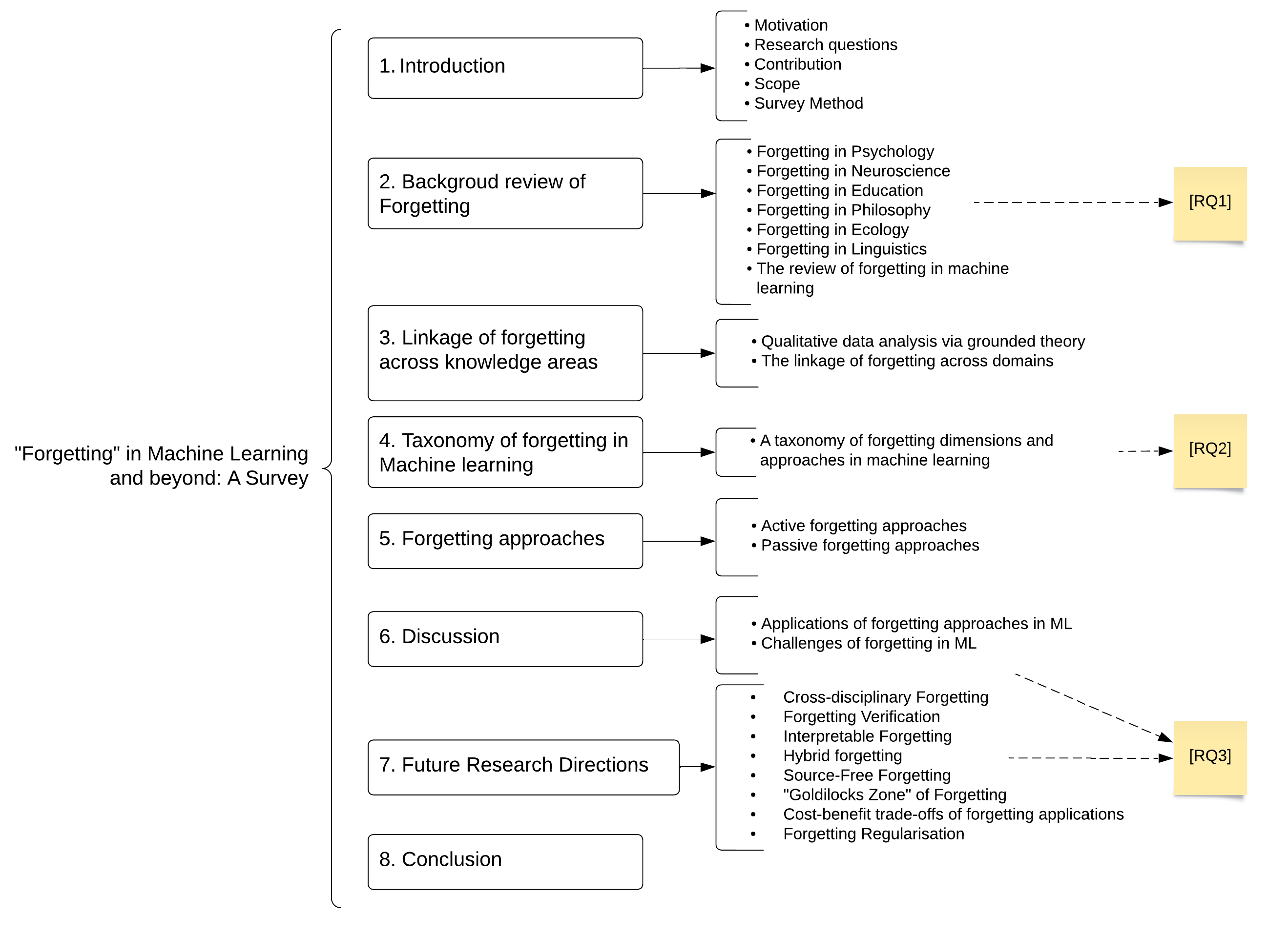}
    \caption{Scope and structure of the survey}
    \label{fig:Scope and structure of the survey}
\end{figure}
In the subsequent sections of this article, we aim to systematically address the aforementioned research questions. Figure \ref{fig:Scope and structure of the survey} presents a conceptual map that serves as a guide to understand the structure and logical flow of our discussion throughout various sections in this survey.

We followed a search strategy to thoroughly retrieve and filter the literature, similar to the method in \cite{Abdelrahman2023}. First, we curated the search queries [\textit{(forgetting OR "memory loss" OR "knowledge decay") AND machine learning}] to match the survey topic and executed them using well-known scientific search engines including Google Scholar, IEEE Xplore, ScienceDirect, SpringerLink, and ACM Digital Library. After collecting around 6,500 articles, we then performed a series of inclusion and exclusion criteria to ensure the relevancy and quality of the collected literature. We filtered out those that were published before 2015 and were not peer-reviewed to focus on recent advances. Subsequently, we further filtered the articles by reading their abstracts to only retain those that discuss the positive side of forgetting in machine learning instead of purely investigating catastrophic forgetting. This reduced the pool to 535 articles. We then sorted the results by multiple factors, including the novelty of the contribution (e.g., proposing a new technique or concept versus extending an existing one), the relevance of the contributions (i.e., the significance of the forgetting approach in the research pipeline), the quality of the publication venue, and the depth of evaluation (e.g., the number and sise of evaluated datasets, performance compared to baselines). Finally, we selected the top 100 articles from the sorted list to be included in this survey. We further performed backward snowballing to find historical methods using references from the included articles and added extra keywords that are not directly related to forgetting but are very related to the topic, such as "machine unlearning," "adaptive learning," and "mitigating negative transfer." This process added approximately 50 additional papers to the set of articles covered by this survey. Data from the selected publications were then extracted using a standardised form, capturing authors, year of publication, research objectives, methodology, key findings, and conclusions.

Following the comprehensive search and filtering of the literature on forgetting in machine learning, we organise the key areas into a structured taxonomy as attributes. This scheme was devised to systematically categorise the various research findings, methodologies, and research gaps uncovered during the review process. Through this taxonomy, we aim to provide a coherent framework that aids in navigating the complexity of the field, fostering a deeper comprehension of how forgetting phenomena are tackled across different machine learning approaches.

The contributions of this research are summarised as follows:
\begin{itemize}
    \item A comprehensive overview of current research in machine learning where the concept of "forgetting" is integrated into model design along with an in-depth discussion of different types of selective forgetting.
    \item Bridge the gap between concepts of forgetting from various knowledge areas and their application in machine learning, deepening our understanding of how mechanisms of human forgetting can inform the design of machine learning models.
    \item Identify the challenges in existing research and ways to enhance model learning by incorporating forgetting, thereby suggesting directions for future research.
    \item Discuss potential biases and ethical issues that may arise with the implementation of forgetting in machine learning, underscoring the importance of responsible research in this area.
\end{itemize}

The remainder of this paper is structured as follows. Section \ref{sec2: background review} reviews the literature on the role of forgetting in various disciplines including computer science, thereby addressing RQ1. In Section \ref{sec3: linkage}, the connections of forgetting across multiple knowledge areas and machine learning are established. Section \ref{sec4: taxonomy} brings up a taxonomy to summarise different types of forgetting dimensions and approaches in machine learning, aiming to answer RQ2. The specific approaches are summarised in Section \ref{sec5: appraoches}. Finally, Section \ref{sec6: Discussion} summarises the applications and challenges of integrating forgetting into machine learning, while also highlighting potential biases and risks, leading up to the exploration of future research directions in Section \ref{sec7: future directions} to address RQ3.

\section{Background of Forgetting in different knowledge areas}
\label{sec2: background review}
This section discusses research that focuses on the beneficial aspects of forgetting such as to decision-making, adaptation, creativity, mental health of human beings across numerous domains \cite{Gravitz2019,Ryan2022}, as well as the generalisation power and evolution of animals \cite{Rodriguez1993,Gagliano2014}. 

\subsection{Forgetting in Psychology: ensure effective memorisation and emotional regulation}
\label{2.1 psychology}
The forgetting curve, proposed by Ebbinghaus \cite{Ebbinghaus2013}, is the most prominent theory in psychology. It shows a rapid decrease in memory retention within the first hour after learning new information, which then slows down over time. Since then, much research has been focused on exploring and adjusting factors to mitigate forgetting \cite{MacLeod2009, Averell2011}. However, forgetting is also posed in psychology as a natural and necessary process that allows humans and animals to prioritise information \cite{Roediger2010,Storm2011}. Jorge Luis Borges illustrated the role of forgetting and memory for the human experience in "Funes the Memorious" \cite{Borges1999}. As Funes is unable to forget anything, leading to a state where the abundance of detail in his memories makes it difficult to generalise, abstract, or prioritise information. The story reflects on the psychological concept of "cognitive overload" \cite{Sweller2011}, where too much information can overwhelm the processing capacity of the working memory, leading to difficulties in comprehension, decision-making, and learning. 

Emotion regulation, the attempts to control when and how people experience and express emotions, also involves selective forgetting \cite{Nrby2018}. 
Such forgetting may allow for a focus on positive memories and thereby help form a mnemonic basis for optimism as well as active and explorative approach behaviour. Relevant studies in clinical psychology also proposed that deficits in active forgetting function may be a causative factor in psychopathologies, like post-traumatic stress disorder, depression, schizophrenia, and obsessive-compulsive disorder \cite{Hertel2003,Costanzi2021}. "Forgetting and reconstruction hypothesis" \cite{Lee1983} suggests that "reloading" previously learned knowledge helps improve long-term learning \cite{Bjork2019}. Human memory is composed by two components: forgetting and learning \cite{Bjork2019,woodward1971}. The forgetting component is about the loss of retrieval strength and the learning component relates to the gain in storage strength. Both of them need to be combined to enable efficient knowledge updates.

In summary, research in psychology emphasises that forgetting is not just a failure of memory, but a natural and necessary process that helps prioritise relevant information, which ensures effective memorisation and potentially contributes to psychological health.

\subsection{Forgetting in Neuroscience: the magic of engram cells}
\label{2.2 Neuroscience}
In neuroscience, forgetting is seen as a form of neuroplasticity that alters engram cell (cells related to memory association \cite{VanderLinden2000}) accessibility in a manner that is sensitive to mismatches between expectations and the environment \cite{Ryan2022}. Animals perceive the world by encoding information through a process of learning, forming memories that enable adaptive behaviour and promote survival \cite{James2007}. However, not all memories are maintained equally, with many forgotten after learning \cite{Davis2017}. Summarised by recent neuroscience research \cite{Ryan2022}, five biological mechanisms of natural forgetting naturally occur in our brain: receptor trafficking; spine instability; inhibition; synapse elimination; and, neurogenesis. The prevalence of forgetting in the healthy brain suggests that it represents a dynamic nature rather than a flaw of normal memory functioning \cite{Richards2017,Nrby2020}. This dynamic nature is driven by the process of expectancy violation \cite{Ryan2022,Gershman2019}. According to this hypothesis, when expectancies are reinforced, or new salient information is linked to an existing memory, positive prediction errors drive reconsolidation and memory updating. When expectations are violated, negative predictions drive the forgetting of seemingly inconsistent information \cite{Miller2021FailuresMemories}. 

Research also shows that forgetting, as a form of adaptive learning, helps us to change and adapt in response to experiences, thus preventing experience overfitting \cite{Gravitz2019}, which refers to the tendency to rely too heavily on past experiences, leading to a failure to adapt to new situations. In the knowledge updating process, where neuroplasticity occurs, forgetting can help the brain reorganise itself and prune unnecessary connections, making space for new learning and memory. These findings suggest that it is helpful to consider learning and forgetting as different aspects of a cognitive process of belief updating, where an organism is constantly changing how to respond to environmental affordances \cite{Pignatelli2019,Ryan2022}.

\subsection{Forgetting in Education: a form to promote long-term learning} \label{Forgetting in Education}
\label{2.3 Education}
In the field of education, forgetting and learning are not opposing forces but interconnected processes that shape a student's educational journey \cite{Bjork2019}. A common misconception equates immediate performance with learning, overlooking that true learning can only be assessed after a delay, whereas performance reflects the short-term recall of knowledge or skills \cite{Soderstrom2015}. This critical distinction, recognised since the 1930s \cite{Tolman1930}, underscores the need to differentiate between short-term performance and long-term learning.

The factors that produce forgetting but enhance learning are driven by a few effects in education. The Context Effect suggests that changing study contexts can introduce desirable difficulties \cite{Smith1978}, improving recall by blending subtopics rather than tackling them sequentially \cite{Bjork2014,Cooper2018}. The Spacing Effect demonstrates that spaced learning, as opposed to cramming, leads to better long-term retention by revisiting material over time \cite{Bjork1970,Teninbaum2016}. Lastly, the Generation Effect shows that actively generating information enhances recall more than passive study \cite{Slamecka1978}, although this effect requires effective feedback to prevent the downsides of forgetting \cite{Kornell2009}. This is consistent with the principles of "flow theory" \cite{Sharek2011}, where the state of flow occurs when the learning content is neither too easy nor too frustrating, allowing learners to fully engage in the learning practice.

The literature in education suggests that forgetting can aid in long-term learning, but it is essential to regulate the extent of forgetting to help students stay in the flow channel.

\subsection{Forgetting in Philosophy: multidimensional self-awareness and moral duty in humanhood} 
\label{2.4 Forgetting in Philosophy}
In philosophy, forgetting is recognised as an essential element of human existence, contributing to the development of personal identity and the construction of self-narratives. As stated by Krondofer \cite{Krondorfer2008}, "Without forgetting, the human species would have to continuously relive the past and never fully experience the present moment. Without forgetting, there would be no future."
%%forget in epistemology
Forgetting plays a crucial role in shaping what we reasonably believe by losing not just evidence but also counter-evidence, making it significant in evaluating epistemological theories \cite{Matthew2018,Dougherty2011,Frise2017}. Normal human forgetting “approximates a virtue” rather than a vice, which represents a balanced approach between excessive forgetting and excessive remembering, thus, underlies a normative epistemic status \cite{Michaelian2011}. Furthermore, forgetting can be associated with a gradual wear of memory connections, which helps shape personal identity \cite{Behan1979}, yet excessive forgetting might undermine it. Thus, understanding the specifics of what and when to forget is crucial for examining these impacts. 

Various memory types, such as semantic, episodic and procedural, bring multiple dimensions of forgetting \cite{Michaelian2011,Tulving1983}. Philosophically, forgetting is seen both as a mental state and a process over time. Theories suggest that forgetting occurs either as permanent memory loss or temporary inaccessibility \cite{Matthew2018,Harris2021}. Matthew's Reductionist Process Theory \cite{Matthew2018} views forgetting as evolving through stages, indicating that the depth of forgetting progresses with time, which is similar to Ebbinghaus's forgetting curve \cite{Ebbinghaus2013}. This concept highlights the dynamic nature of forgetting, emphasising the importance of time and gradual change in understanding memory loss, rather than viewing it as a static state. Furthermore, the LEAD theory \cite{Matthew2018} suggests that forgetting and relearning are interconnected, allowing for knowledge retention alongside memory loss \cite{Michaelian2016,Bernecker2008}. It views memory as an active process that reconstructs past information \cite{Carava2021}, where difficulty accessing memories is seen more as a generative challenge than mere storage failure. 

Feedback mechanisms unconsciously regulate memory retrieval, similar to control loops in engineering systems that adapt to changing operational conditions. This generates a metacognitive feeling that acts as feedback on our cognitive process \cite{Arango-Munoz2013,Arango-Munoz2014}, serving as evidence of forgetting in philosophical views \cite{Halamish2011,Michaelian2016}.

Just as forgetting influences our beliefs, it also allows us to manage aspects of our identity and facilitate the process of letting go \cite{Bernecker2018,Murray2019,Broome2013,Williamson2002}. \cite{Basu2022} introduced "Siloing forgetting", which involves segregating undesirable memories. 
The "Fruit of the Poisonous Tree"\footnote{The phrase, “fruit of the poisonous tree,” was first used in 1939 in Nardone v. United States, 308 U.S. 338.} theory, from a moral perspective, discusses the importance of not only believing what is true but also obtaining truth in the appropriate manner \cite{Basu2022}. Similarly, what should be forgotten is often determined not only by the content of the information but also by how it was acquired in the first place \cite{Basu2022,Munch}.

In summary, research about forgetting in philosophy suggests the existence of multiple forgetting dimensions. 
Moreover, the discussions on moral duties of forgetting inspired us to consider the ethical manner of the collection and use of data for training machine learning models.
\subsection{Forgetting in Ecology: trade-offs in animal behaviour and plant habituation}
\label{2.5 Ecology}
Forgetting plays a pivotal role in striking a balance between flexibility and the associated costs incurred by both animals and plants in their behavioural adaptations \cite{Tello-Ramos2019}.
Psychologists have defined cognitive flexibility as the ability to adjust and reverse contingencies while acquiring new information \cite{Badre2006}, contributing to enhanced survival and reproductive success \cite{Snell-Rood2013,Dukas2004,Lefebvre2004}. However, cognitive flexibility is not a random occurrence; it is influenced by environmental conditions, affecting animals' learning and forgetting rates. In places with seasonal food changes, animals develop better spatial memory to survive \cite{Hampton1998}, known as the harsh environment hypothesis \cite{Hermer2021,Sonnenberg2019,Pravosudov2002}. 
A potential cost of being cognitively flexible is handling proactive interference, due to the cue-overload principle \cite{Watkins1976}. Experiments with mice suggest that this interference can be reduced by distinctly varying the context of tasks \cite{Rodriguez1993}, similar to the "context effect" \cite{Smith1978}. 

Similar to animals, plants also adapt to changing environments through a form of learning and forgetting, as seen in Mimosa pudica's leaf-folding habituation to repeated disturbances, indicating their capability to process and respond to stimuli \cite{Gagliano2014}. These adaptive responses are essential for their survival, allowing plants to selectively remember and utilise information to adjust their behavior in response to external cues \cite{Hemmi2009,Giles2009,Eisenstein2001,Thorpe1956}. Nevertheless, plants' development of adaptive abilities involves a trade-off between energy expenditure and survival benefits, such as balancing the predation risk (displaying visible leaves) and energy gain (opportunity for photosynthesis with open leaves) \cite{Jensen2011}.

Research in ecology suggests the existence of trade-offs: between memory retention and the acquisition of new memories in animals, and between energy allocation for survival and growth in plants. Forgetting can be seen as a mechanism allowing these trade-offs, thus, increasing the adaptability of some animals and plants. 
\subsection{Forgetting in Linguistics: shaping language evolution and historical narratives} 
\label{2.6 Forgetting in Linguistics}
\emph{"Memories are crafted by oblivion as the outlines of the shore are created by the sea."} One of Auge's \cite{Auge2004} most memorable portraits vividly implies the power of forgetting. Forgetting has been a subject of extensive investigation within the linguistics, specifically focusing on the evolution of language itself and the forgetting of its application.
%对于语言本身的遗忘：语言减损与语言生成 -> 适应性和延续性/创造性

Previous research has identified seven theories of forgetting in language attrition: repression/suppression, distortion, decay, interference, retrieval failure, cue dependency and dynamic systems theory \cite{Ecke2004,Kopke2007}. These theories shed light on the underlying regularity and rationality of the forgetting process in language attrition. Similar to language acquisition, speech can also be forgotten. Daniel Heller-Roazen's "Echolalias" \cite{Heller-Roazen2005} examines the complex nature of linguistic forgetfulness, where language forgetting often reflects demographic shifts and changes in expression. A prime example is the extinction of the Ubykh language in 1992, marked by the death of its last native speaker \cite{Nettle2000}. At the meanwhile, the end of one language can lead to the birth of new forms of expression and creativity, both communally and individually, as illustrated through Jakobson's Regression Hypothesis and the transformation of Hebrew in Jewish scriptures \cite{Keijzer2004,Jakobson1941, Heller-Roazen2005}. This process underscores the essence of the speaking animal who, in the act of forgetting, discovers new realms of inspiration.

%对于使用语言方式的遗忘：神学；修辞学 （公共政治与未来主义）；感情（爱与原谅）；历史观
In linguistic discussions, the concept of forgetting also has implications for our understanding of history, politics and lifestyle. Early thinkers such as Plato \cite{Plato1962}, Augustine \cite{Augustine1991} and Shakespeare \cite{Shakespeare2019} viewed forgetting as a negative phenomenon, representing a loss of wisdom. However, contemporary perspectives acknowledge that forgetting can have positive effects on shaping public memory and rhetoric, offering political and moral advantages \cite{Margalit2002,Volf2021,Ricoeur2004,Auge2004}. The process of collective forgetting, which refers to the public's selective memory, influences individuals' interpretations of the past and future \cite{Margalit2002}. Additionally, discussions on forgetting prompt considerations regarding the political biases inherent in training machine learning models and the potential danger signals they may carry \cite{Muller2022}.

The merit of forgetting is also linked to forgiveness in linguistics, suggesting that forgiveness comes from intentional and redemptive remembering \cite{Hawhee2009}. This process, seen as a form of emotional expression, bridges confession and forgiveness, promoting spiritual equality as per Kierkegaard's concept of "true love" \cite{Volf2021,Kierkegaard1946}. Additionally, different dimensions of forgetting have been proposed in rhetoric, namely, Return; Suspense; and Rebeginning, with each representing a particular temporal duration \cite{Auge2004}.

In conclusion, the research about forgetting language itself in linguistics showcases the continuity and creativity of expression. And the forgetting of language use, at the same time, shapes our perspectives on history, politics and the emotion of forgiveness.

\subsection{The review of forgetting in machine learning}
Forgetting in machine learning refers to the intentional process of removing specific data or influences from a model's training history, often to update its knowledge for better performance \cite{Zhou2022}, comply with privacy requirements \cite{Cao2015} or correct biases \cite{Peng2021}. There are review studies that summarise machine learning approaches associated with "forgetting" in several different ways. For example, \cite{Nguyen2022,Zhang2023} summarise forgetting approaches in machine learning under the machine unlearning domain, which is driven by user privacy concerns; \cite{Muller2022} provides a comprehensive analysis of how forgetting has been applied in HCI studies; \cite{Wang2023} reviews forgetting approaches from the deep learning perspective, while others touch on "forgetting" indirectly by summarising downstream machine learning tasks that are highly relevant to forgetting \cite{Zhang2023a, Pan2009,Mehrabi2021}.

However, these existing reviews on forgetting in ML fall short in the following gaps: 1) The discussion from many of them only stays on the technical level \cite{Nguyen2022,Wang2023}, thus lacking the linkage with forgetting theories embedded in other disciplines. 2) Some machine learning reviews only cover forgetting approaches for one narrow problem \cite{Zhang2023,Zhang2023a,Muller2022}, which misses a complete perspective to summarise the findings.
To address the above research gaps, this survey aims to establish the linkage from forgetting methodologies in machine learning to forgetting theories in other domains and propose a comprehensive taxonomy for a more complete summary of forgetting involving multiple ML tasks.

\section{Expand The linkage of forgetting across knowledge areas to machine learning}
\label{sec3: linkage}
\begin{figure}[h]
    \centering
    \includegraphics[width=0.5\linewidth]{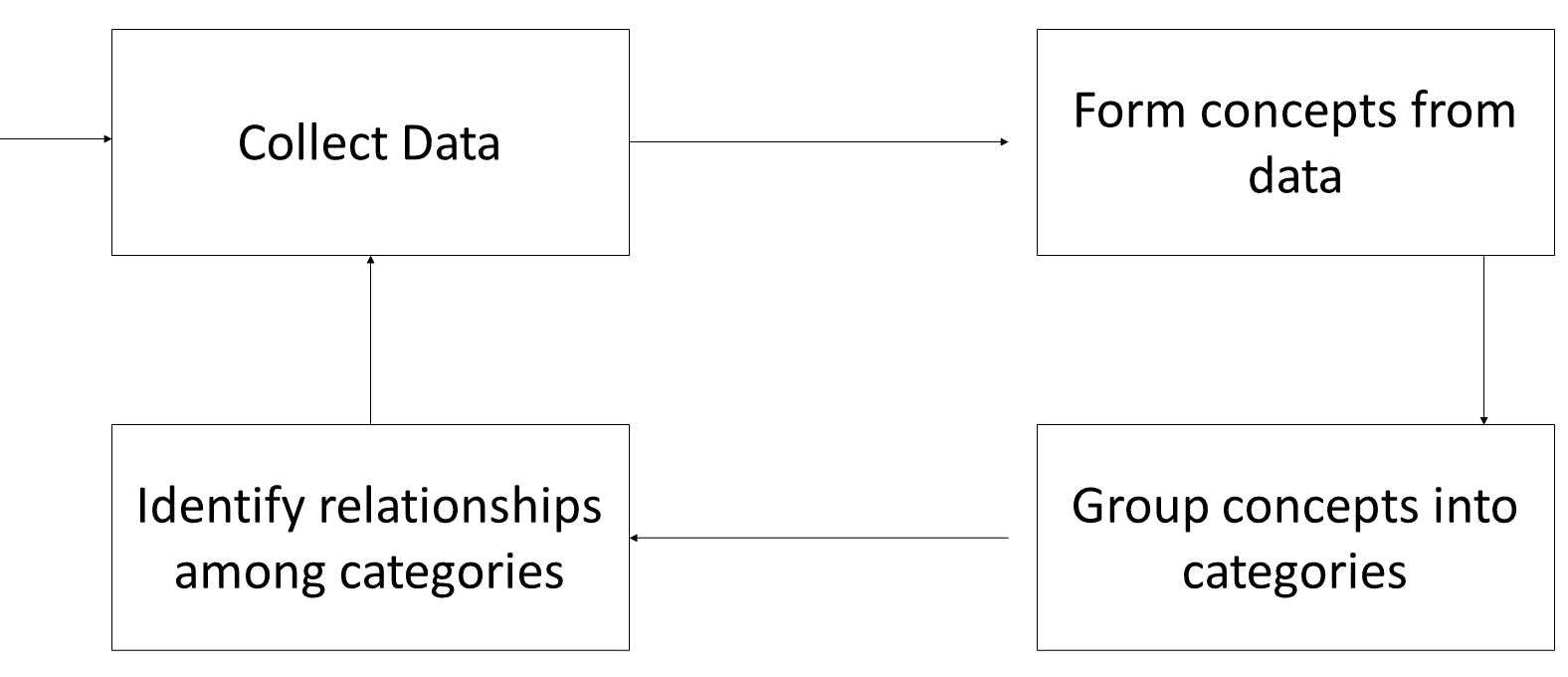}
    \caption{The workflow of qualitative data analysis using grounded theory}
    \label{fig:grounded theory workflow}
\end{figure}

%\subsection{Qualitative analysis via grounded theory}
We adopted a qualitative data analysis approach rooted in the grounded theory to helps us to extract insights about forgetting across multiple domains systematically and build the linkage among them. %integrating them into a coherent framework \cite{Corbin1990}. 
The principles of grounded theory enable us to explore a range of themes and patterns in the data, identify underlying relationships among entities, and categorise them into a theoretical framework that explains the forgetting phenomena \cite{Corbin1990}, which helps to ensure the research is grounded in the data and thus can generate valid and reliable findings \cite{Walker2006, Noble2016}. Following the grounded theory, our qualitative data analysis is broken down into three stages: open coding, axial coding and selective coding \cite{Oktay2012}. The overall workflow of qualitative data analysis using grounded theory is shown in Figure \ref{fig:grounded theory workflow}.
%These stages are designed to refine the understanding of data progressively, moving from identifying key points to integrating them into a coherent framework \cite{Corbin1990}. 

We began by gathering qualitative data (peer-reviewed research papers) on how forgetting occurs in fields such as cognitive psychology, neuroscience, and linguistics \cite{Noble2016,Dickersin1994}. During the open coding stage, we identify and label concepts, debates, and themes within the papers, breaking down the data into discrete parts for comparison. In the axial coding stage, we reassemble the data by making connections among the fragmented ideas, building categories that relate the labelled codes through a mix of inductive and deductive reasoning. This process reveals patterns and similar views across different fields, such as the emphasis on feedback signals in both neuroscience and philosophy \cite{Arango-Munoz2013,Ryan2022}, converging on themes like the generative power and adaptability of memory systems \cite{Gravitz2019,Carava2021}. During the selective coding stage, we integrate and refine these categories to construct coherent themes representing our findings on forgetting. For example, we identify topics such as the dimensions of forgetting and inherent biases are frequently discussed across disciplines during this stage. We continuously incorporate new data to refine our framework, iterating until we achieve a comprehensive understanding of forgetting from various perspectives.

\begin{figure}
    \centering
    \includegraphics[width=1\linewidth]{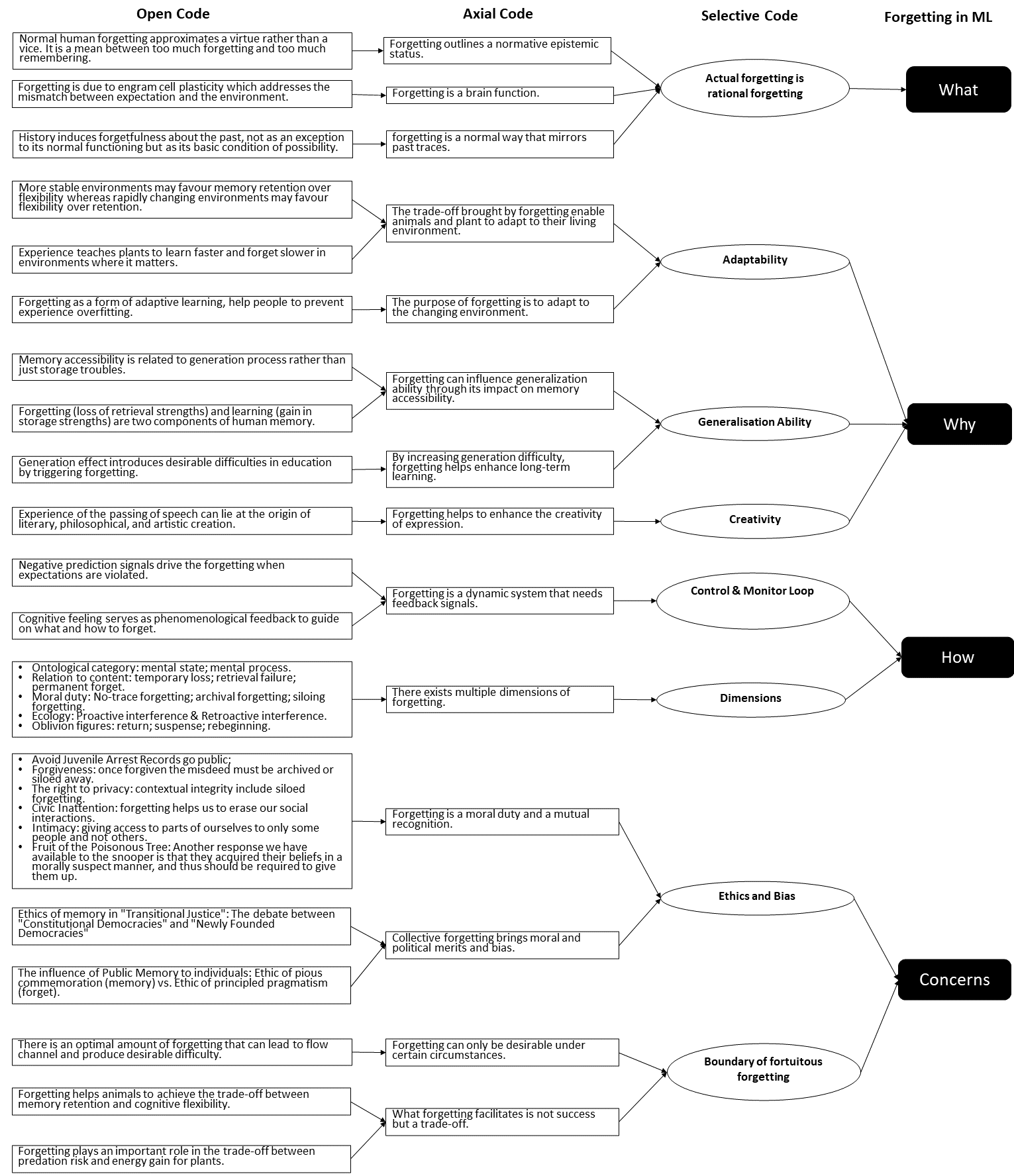}
    \caption{Quantitative analysis on forgetting across different domains using grounded theory}
    \label{fig:grounded theory}
\end{figure}

Building on selective coding, we extend the key categories of research on forgetting from other domains to the field of machine learning, as presented in Figure \ref{fig:grounded theory}. The highlights from background review have been categorised into four sections below to expand our discussion, which inspired us to establish linkage and develop more concrete ideas in the exploration of the role of forgetting in machine learning. 

\subsection{Actual forgetting is rational forgetting}
The first common denominator in the literature is the assessment of the phenomenon of the presence of forgetting. Hegel's famous statement \cite{Hegel1998}, "What is actual is rational", reflects the recognition of forgetting as a significant aspect of human cognition. Rather than being seen as a flaw, forgetting is viewed as a legitimate and normal functionality. Neuroscientific research supports this perspective by attributing forgetting to the natural process of engram cell plasticity in the brain \cite{Ryan2022}. Philosophical theories also contribute to our understanding of forgetting, presenting it as a mean to shape personal identity and establish a normative epistemic status \cite{Matthew2018}. Additionally, in rhetoric, forgetting is recognised as a normal process that allows us to mirror past traces and serves as a fundamental condition for constructing historical narratives \cite{Hawhee2009}. Considering these insights, it becomes evident that forgetting in computational models must also have its place. This hypothesis encourages us to take a positive stance on the important role of forgetting not only in human learning but also in machine learning.

\subsection{Unlocking Potential: The Role of Forgetting in Adaptability, Generalisation, and Creativity} 
The rationality of forgetting lies in its positive impact on the sustainability of species, both physiologically and psychologically, particularly regarding its contributions to adaptability, generalisation, and creativity.

Adaptability is emphasised in animal behaviour research, where forgetting is proven that can help decrease proactive interference and thus help animals to make flexible reactions to the changing environment according to the cue-overload principle \cite{Tello-Ramos2019}. Similarly, plants also exhibit forgetting behaviour, which aids in the storage and retrieval of biological information, allowing them to provide adaptive responses to their surroundings \cite{Gagliano2014}. The same purpose is proposed in neuroscience as well, where forgetting, as a form of adaptive learning, prevents individuals from experiencing overfitting \cite{Gravitz2019}. Notably, the challenge of overfitting is not limited to natural organisms but also pertains to machine learning. The observations from ecology and neuroscience highlight the broader significance of forgetting in promoting adaptability and addressing challenges in different domains, contributing to the sustainability and efficiency of computational systems.

%Generalisation ability -strategies to enhance learning
The facilitation of generalisation ability and creativity through forgetting is emphasised across various disciplines. In cognitive psychology, forgetting is characterised as being linked to retrieval rather than storage capabilities \cite{Bjork2020DesirablePractice}. From educational perspectives, a certain degree of forgetting can compel us to form new connections between new cues and existing memories, thereby enhancing generalisation and summarisation abilities \cite{Bjork2019}. Moreover, at an individual level, the forgetting of speech can serve as a catalyst for literary, philosophical, and artistic creations from linguistic perspective \cite{Heller-Roazen2005}. Leveraging the concept of forgetting may thus enable the development of more advanced generalisation and innovation capabilities for machine learning.

\subsection{How the process of forgetting is shaped?}
\label{2.7.3How the process of forgetting is shaped?}
%Control & Monitor loop
Once the benefits of forgetting have been elucidated, various disciplines have offered insights into effective strategies for forgetting. One crucial approach is the establishment of feedback mechanisms or control and monitoring loops. Neuroscience research emphasizes that forgetting is a dynamic system that relies on feedback signals, which represent our expectations of the environment \cite{Ryan2022}. In philosophy, these signals manifest as cognitive feelings, serving as phenomenological feedback that guides us in what and how to forget \cite{Arango-Munoz2013,Arango-Munoz2014}.

%Dimensions of forgetting 
The process of forgetting is intricately linked to different dimensions, as emphasised in philosophy and linguistics. Philosophical theories classify forgetting based on its temporal dimensions, distinguishing between states and processes \cite{Matthew2018}. Moreover, forgetting can be categorised based on the way it occurs, encompassing temporary loss, retrieval failure, or permanent forgetting \cite{Michaelian2011,Tulving1983}. Additionally, the dimensions of forgetting intersect with moral duty, giving rise to classifications such as No-trace forgetting, archival forgetting, and siloing forgetting \cite{Basu2022}. The definition and classification of these diverse dimensions of forgetting inspire us to contemplate the myriad modes of archiving within the field of machine learning and the consequential implications they entail.

\subsection{Challenges in forgetting}
\label{background:forgetting boundary}
%Ethics and Bias 
Although our focus primarily revolves around the benefits of forgetting, it is crucial to acknowledge the accompanying issues and challenges. One notable concern arises from the ethical implications discussed in philosophy and linguistics. Forgetting, as a reflection of remembrance, significantly shapes our perception of ethics, moral duty, history, politics, and the past and future \cite{Basu2022,Matthew2018}. Given this influence, it becomes imperative to exercise caution when dealing with the forgetting behavior of machine learning systems. Different types of data and their weighting during model training can impact the reasoning and decision-making abilities of the model. Hence, it is essential to carefully consider the ethical concerns and biases associated with forgetting in this context.
%Boundary of forgetting
Another significant aspect to consider is the boundary of fortuitous forgetting, as not all instances of forgetting yield positive outcomes. In educational theories, excessive forgetting can render problem-solving overly challenging, potentially blocking students' flow channel of learning thus creating negative experiences \cite{Vann2020}. As mentioned before, forgetting is a method for animals to achieve trade-off between memory retention and cognitive flexibility to adapt to changing environments. However, excessive forgetfulness in a stable environment can lead animals to forget vital information such as their homes or the location of stored food, making them vulnerable to predation \cite{Tello-Ramos2019}. Similarly, unfit archival practices may result in catastrophic forgetting and nonsensical decision-making for machine learning models, which presents us with the new challenge of determining suitable boundaries for forgetting in different types of training tasks, ensuring that forgetting is employed effectively and appropriately.

\section{A taxonomy of forgetting in machine learning}
\label{sec4: taxonomy}
\begin{figure}
    \centering
    \includegraphics[width=1\linewidth]{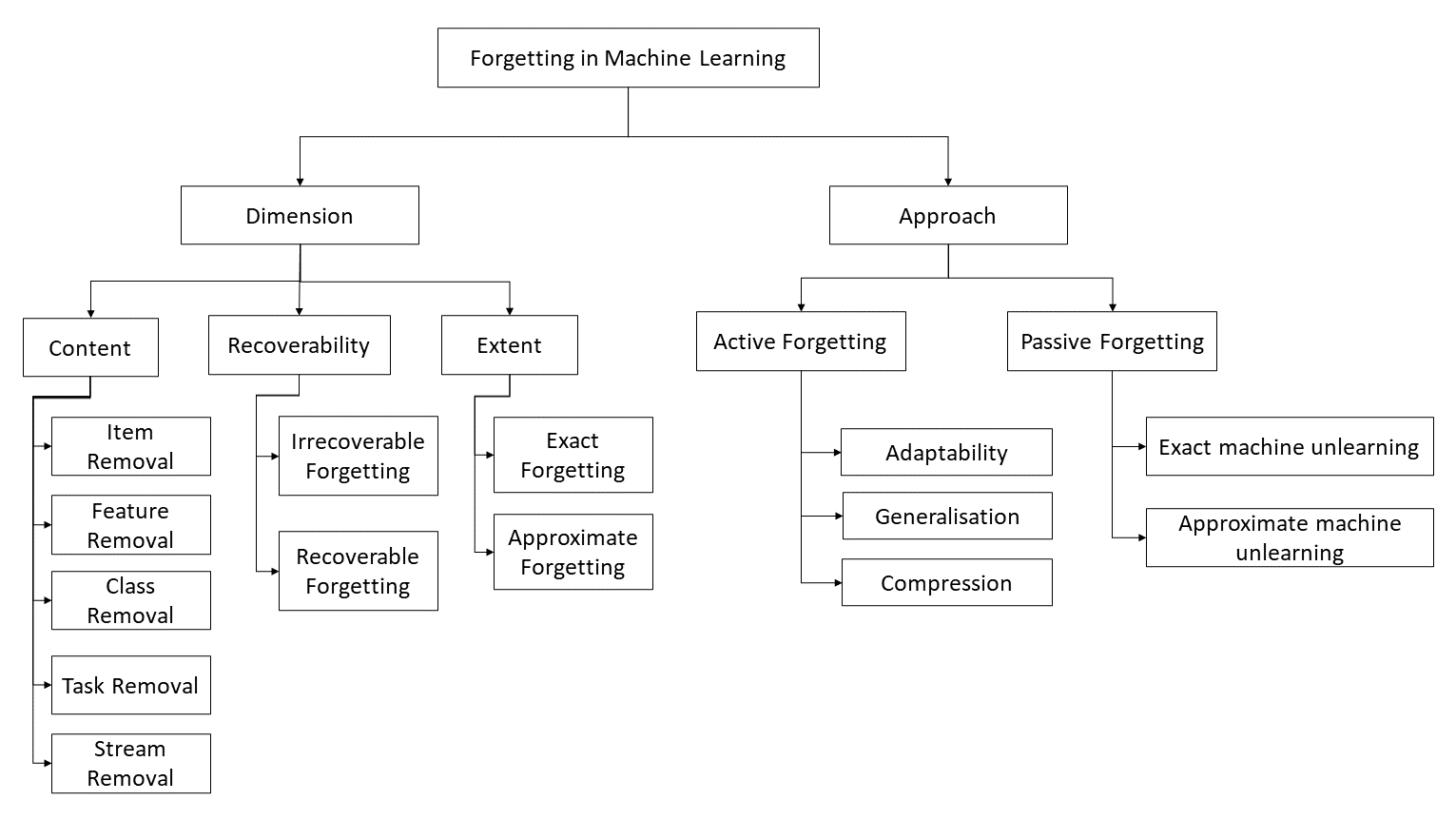}
    \caption{A taxonomy of forgetting in machine learning}
    \label{fig:taxonomy}
\end{figure}

Built upon the expanded linkage from forgetting in other knowledge areas, we present a taxonomy (Figure \ref{fig:taxonomy}) with the main aspects of forgetting in machine learning. This taxonomy serves as a framework for understanding the different dimensions and approaches through which machine learning models can 'forget' or 'unlearn' information. This exploration will not only clarify the concepts but also highlight the practical implications and challenges associated with each. By doing so, we aim to provide clarity and structure to the complex process of forgetting in machine learning, setting the stage for further investigation and application in this field.
%The remainder of this section will delve into the components of this taxonomy, articulating their meanings and interrelationships.

\subsection{The dimensions of forgetting in Machine Learning}
Drawing upon prior research from different domains, this paper summarise the dimensions of forgetting in machine learning into three categories, delineated by the content, recoverability, and extent of forgetting as shown in Figure \ref{fig:taxonomy}.

\subsubsection{Content of forgetting}
The content of forgetting in machine learning models is influenced by various removal requests, which in turn guide the strategies for forgetting. The types of content that can be removed span multiple levels, each representing a distinct dimension of forgetting. \textbf{Item-level Forgetting} is concerned with the exclusion of specific data instances from the training dataset, addressing the most common requests for forgetting. Removal requests may arise not just for individual data items, but for clusters of data sharing common features or labels, leading to \textbf{feature-level forgetting} \cite{Warnecke2021}. This is particularly critical when data attributes become outdated or misleading. 
\textbf{Class-level forgetting} encompasses the targeted removal of entire classes from the model, a process that can be intricate due to computational requirements and its potential effects on the model's efficacy \cite{Tanha2020}. For example, in face recognition systems, each class represents an individual's facial data, and opting out necessitates the elimination of that data class.
\textbf{Task-level forgetting} is essential in continual or lifelong learning, where there might be requests to remove data linked to a particular task. Consider a robot trained for home patient care which must later discard that task from its memory after the patient's recovery \cite{Liu2022a}.
Lastly, \textbf{Stream-level forgetting} addresses the challenge of handling vast volumes of online data with the constraint of limited storage \cite{Nguyen2017}, which involves managing a sequence of data removal requests.

\subsubsection{Recoverability of forgetting}
The categorisation of dimensions based on approaches to forgetting finds its origins in the field of philosophy, where forgetting is classified into distinct dimensions such as archival forgetting, no-trace forgetting, and siloing forgetting \cite{Basu2022}. Each of these categories is associated with unique moral duties, as discussed in \ref{2.7.3How the process of forgetting is shaped?}. In the context of machine learning, a similar categorisation of dimensions can be observed concerning the approaches to processing forgetting. These dimensions are typically distinguished as recoverable forgetting and irrecoverable forgetting. 

\textbf{Irrecoverable forgetting}, often referred to as non-trace forgetting, involves the deliberate removal or deletion of specific knowledge stored within the model. This action is typically undertaken for reasons related to privacy or copyright concerns. It is essential to note that once this knowledge is expunged, it cannot be recovered or restored. The elimination of extraneous or irrelevant information further enhances the scalability of transformers, enabling them to allocate resources more efficiently to attend to a broader array of memories \cite{Sukhbaatar2021}.

On the other hand, \textbf{recoverable forgetting}, also known as archival forgetting, pertains to a scenario in which temporarily forgotten information or knowledge can be restored or retrieved through appropriate methods or processes. In essence, there exists a means to recover the previously forgotten information. This approach explicitly allows for the extracted knowledge from pre-trained models to be temporarily set aside and subsequently reintroduced into the model whenever required. The overarching objective is to equip models with a flexible learning strategy that encompasses forgetting as well. This flexibility grants users a high degree of control over task-specific or sample-specific knowledge while simultaneously ensuring intellectual property protection, as discussed by \cite{Ye2022}.
Recoverable forgetting encompasses several subdomains including ephemeral, inhibitory and gradient forgetting. Ephemeral Forgetting pertains to transient or temporary forgetting, where the model's memory of specific items diminishes over time but can potentially be restored. Inhibitory Forgetting involves situations where the model puts a block on certain memories on purpose, making it hard to recall those specific details unless the block is removed. Lastly, gradient forgetting refers to the model forgetting different things at different rates because of how it learns or optimises its knowledge which could be adjusted.

\subsubsection{Extent of forgetting}
\label{subsection: Extent of forgetting}
The dimensions of machine forgetting can be delineated into two categories based on the extent of forgetting: exact forgetting and approximate forgetting. \textbf{Exact forgetting} entails that the model's outputs, after the removal of a sample, remain the same as those of a model that was never trained on the removed sample. In essence, after exact forgetting, not only is the specified data eliminated, but also its influence on the model is entirely eradicated. From a model-building standpoint, a reasonable criterion for exact forgetting is that the system's state is adjusted to replicate the state it would have in the complete absence of the forgotten data \cite{Yan2022}.

Nevertheless, \textbf{approximate forgetting} involves adjustments to the model and dataset that eliminate the need to retrain from scratch. This forgetting method entails diminishing the influence or significance of specific information within a machine learning model, rather than erasing it entirely \cite{Izzo2021}. Within this approach, the model retains a downscaled or less potent version of the forgotten knowledge. The boundaries and extent of forgetting are less stringent in approximate forgetting as compared to exact forgetting. Approximate forgetting is often used when one wants to maintain a certain level of adaptability or the ability to revisit and fine-tune previously learned information while reducing its current influence on the model.
\subsection{The types of forgetting approaches in Machine Learning}

\begin{figure}
    \centering
    \includegraphics[width=0.75\linewidth]{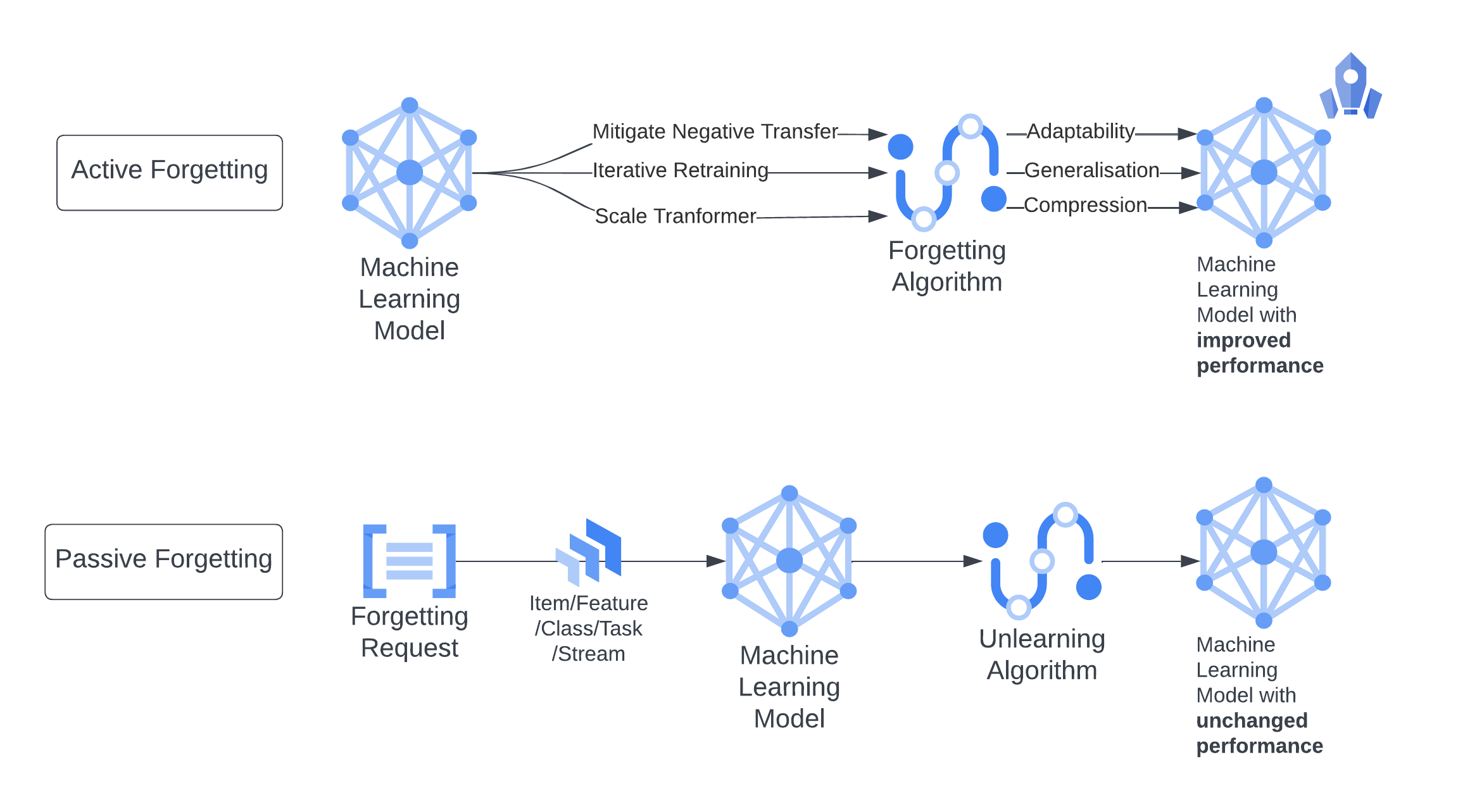}
    \caption{The workflow of active and passive forgetting approaches}
    \label{fig:active passive workflow}
\end{figure}

This survey summarise the forgetting approaches in machine learning into two branches based on their motivations: active and passive forgetting approaches. The first branch, where machine learning models are designed to actively perform forgetting with the primary goal of updating information to improve performance, is classified as active forgetting approaches. The second branch, where forgetting occurs only in response to specific requests to protect user privacy and enhance security, is categorised as passive forgetting approaches. The different workflows of each are shown in Figure \ref{fig:active passive workflow}. To provide clearer narratives, the following sections will follow the taxonomy to summarise the categories of active forgetting approaches and passive forgetting categories separately.
% 4 subsubsub active forgetting section
%---------------------------modify from here-----
\subsubsection{Active forgetting}
%三段总结
%1. adaptability
One purpose of using an active forgetting approach to improve model performance is to \textbf{enhance the adaptability} of models when switching tasks. This requires the models to continue performing well on different or unseen data. This is a central task in transfer learning, which enhances machine learning models by utilising data from different domains, aiming to overcome the limitations due to disparities between training and testing data \cite{Pan2009}. However, this approach faces the challenge of Negative Transfer (NT), where leveraging data from the source domain negatively impacts performance in the target domain \cite{Rosenstein2005}. This issue, which is also rooted in educational psychology \cite{Chen1989,Watson1938,Chen2020a}, underscores the importance of carefully selecting related source data to avoid detrimental effects on learning. Recent research has focused on the concept of 'forgetting' irrelevant information as a strategy to mitigate NT and improve knowledge transfer \cite{Wang2019b, Zhang2023a, Xie2017}, indicating a promising research direction. This involves aligning TL techniques with broader theories of forgetting to improve model adaptability. Such approaches include estimating knowledge domain similarity, enhancing data and model transferability, and refining the training process to mitigate negative transfer.

%2. generalisation
One objective for ML models is not just to perform well on the training data but to \textbf{develop generalisation ability} and apply that to unseen data. The ideal model with generalisation ability would extract the underlying patterns or concepts that are universally true, rather than memorising specific details or statistical quirks in the training data. Active forgetting approach plays a critical role in achieving such optimal generalisation. More information does not equal to more learning. Instead of retaining all information, a model can employ selective forgetting to discard noise and irrelevant details. This strategy helps prevent the model from overfitting to the training data by "forgetting" information that limits its understanding of the problem space. Employing an iterative process of forgetting and relearning enables the model to refine its focus on significant patterns and relationships, thereby improving its performance on a variety of unfamiliar datasets.

%3. compression
When comparing human memory capacity to that of machines, it is often argued that humans forget due to intrinsic limitations in brain capacity, whereas machines, equipped with much larger memory capacities, supposedly have no need to forget \cite{Korteling2021,Moravec1988}. This viewpoint, however, simplifies the complex reality. Given that computational resources for ML models are finite, it becomes crucial for them to forget in order to allocate resources more efficiently. This is particularly true for Transformer architectures, such as those used in the GPT series, BERT, and other NLP models, which, despite their high performance across tasks, face significant memory constraints. Their quadratic complexity in relation to sequence length, due to the self-attention mechanism, makes processing long sequences challenging without segmenting inputs or employing approximations. Considering the attention context as a form of random-access memory where each token occupies a slot, the memory size and associated overhead increase linearly with sequence length. To overcome these memory challenges in Transformer architectures, active forgetting emerges as a vital strategy. It allows for scaling transformers to achieve \textbf{lossless compression} through improved attention efficiency and strategies to reduce memory length.

\subsubsection{Passive forgetting}
Different from active forgetting which is performance-driven, passive forgetting process is triggered by removal requests due to data privacy, security or usability concerns \cite{Nguyen2022}. To comply with a removal request, the computer system must eliminate all user data and 'forget' any impact on the models that were trained using that data. Completely eliminating the impact of data marked for deletion is complex. Beyond just removing it from the databases, its impact on artefacts like trained machine learning models must also be nullified. Machine unlearning, as a passive forgetting approach, has been introduced to address the data removal challenge \cite{Zhang2023}. Google's first machine unlearning challenge highlighted that the ideal unlearning algorithm should not only eradicate the influence of a specific dataset but also retain the model's performance \cite{Pedregosa}. 

The unlearning problem is firstly formalised as a game between two entities, the service provider and the user population by Bourtoule et al \cite{Bourtoule2021}. The service provider \( S \) can be an organisation that can collect various users’ information and the collected information are stored in the form of a dataset \( D \). The service provider \( S \) uses these data to train a machine learning model through \( A(\cdot) \) as a way to provide an intelligent service to the user \( U \). Then, according to the GDPR \cite{Mantelero2013}, any user \( u \) in \( U \) has the right to request the removal of part of the data \( D_f \) from \( D \), and the service provider \( S \) must execute it. Thus, the service provider \( S \) must modify the model parameters \( w = A(D) \) to generate a new model with parameters \( w_u = U(\cdot) \), which represents a model without trained data \( D_f \). 

\begin{figure}
    \centering
    \includegraphics[width=0.75\linewidth]{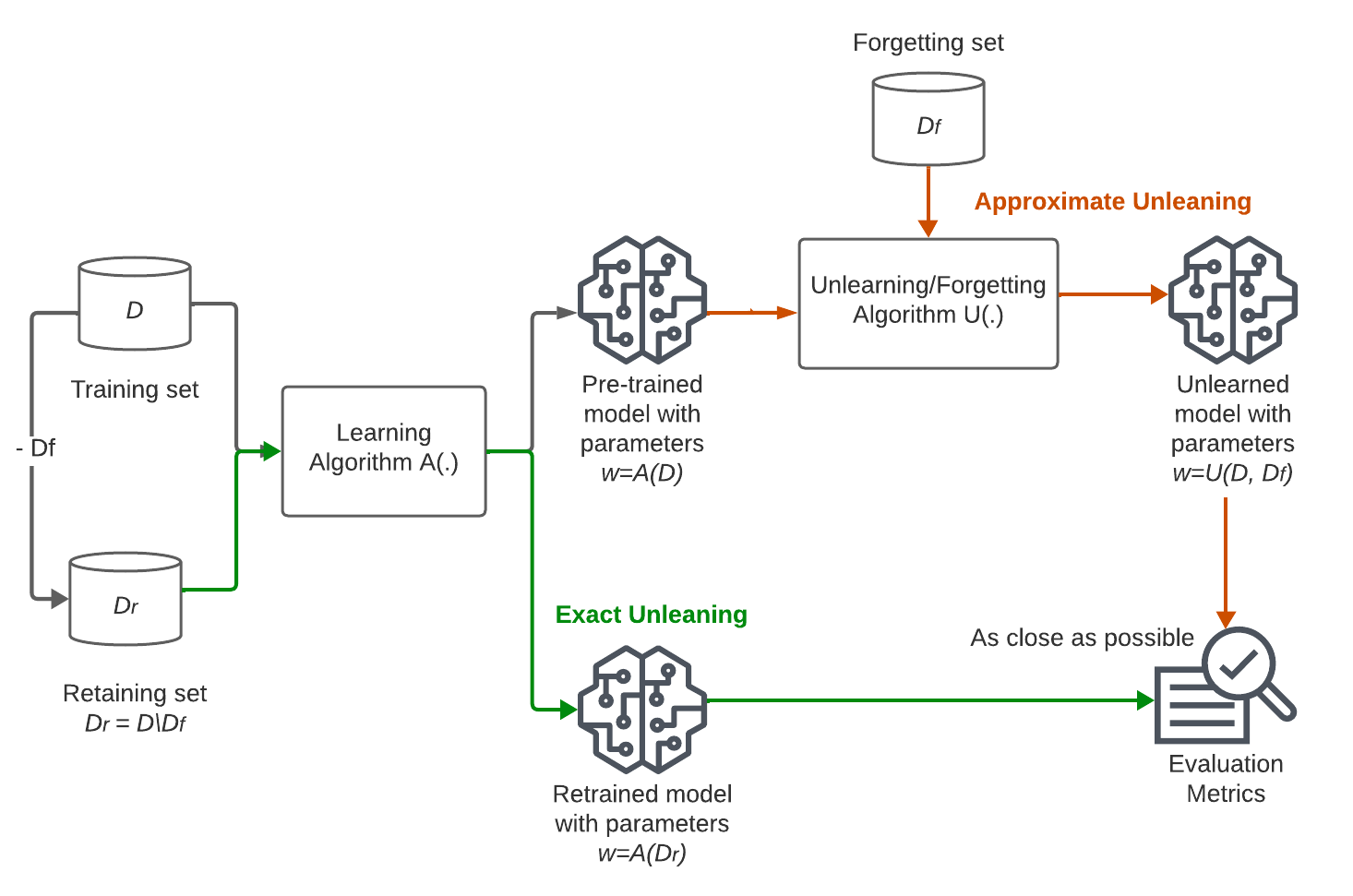}
    \caption{Machine Unlearning Pipeline: An unlearning algorithm takes as input a pre-trained model and one or more samples from the train set to unlearn (the "forget set"). From the model, forget set, and retain set, the unlearning algorithm produces an updated model. An ideal unlearning algorithm produces a model that is indistinguishable from the model trained without the forget set.}
    \label{fig:unlearning pipeline}
\end{figure}

Currently, there are two strategies to perform unlearning on machine learning models \cite{Zhang2023,Nguyen2022,Wang2023}. 
A straightforward way to produce this unlearned model is to retrain the model on an adjusted training set that excludes the samples from the forget set, commonly termed "\textbf{exact unlearning}" (complete, perfect) \cite{Nguyen2022}. However, this is not always a viable option, as retraining deep models can be computationally expensive. An ideal unlearning algorithm would instead use the already-trained model as a starting point and efficiently make adjustments to remove the influence of the requested data. Therefore, a more efficient way is to modify the machine learning model and dataset to achieve an "\textbf{approximate unlearning}" (certified, bounded) effect \cite{Pedregosa}. Current definitions of machine unlearning seek to make the output of approximate unlearning as close as possible to the output of exact unlearning \cite{Zhang2023,Pedregosa}. The differences between exact unlearning (green line) and approximate unlearning (orange line) are presented in Figure \ref{fig:unlearning pipeline} which consists of the primary stages of machine unlearning. 

\section{Forgetting Approaches}
\label{sec5: appraoches}
This section summarises specific existing approaches following the taxonomy proposed in section \ref{sec4: taxonomy}. The approaches are mapped into the approach categories and the dimensions we proposed, as shown in Table \ref{tab:forgetting_landscape}. Note that many approaches could be applied to various dimensions depending on the task requirement and operating ways. Here, we only tick dimensions that it most possibly could be applied to. It only serves as an overview for readers to navigate literature more easily. 

\setlength\rotFPtop{0pt plus 1fil}
\begin{sidewaystable}
\centering % To center the table
\scriptsize % You may try \tiny if \scriptsize is still too large
\setlength{\tabcolsep}{3pt} % Smaller column padding
\resizebox{\textwidth}{!}{% Resizing the table to fit within the text width
% \usepackage{multirow}
\begin{tabular}{l|l|ll|cllllcllcl}
\toprule
\multicolumn{1}{c}{}                                           &                                                           & \multicolumn{1}{c}{}                                        & \multicolumn{1}{c}{}                                        & \multicolumn{5}{c}{\textbf{Content of forgetting}}                                                                                              & \multicolumn{2}{c}{\textbf{Recoverability of   forgetting}}             & \multicolumn{2}{c}{\textbf{Extent of forgetting}}                       & \multicolumn{1}{c}{}                                                                                                                          \\
\multicolumn{1}{c}{\multirow{-2}{*}{\textbf{Taxonomy-parent}}} & \multirow{-2}{*}{\textbf{Taxonomy-node}}                  & \multicolumn{1}{c}{\multirow{-2}{*}{\textbf{Approach}}}     & \multicolumn{1}{c}{\multirow{-2}{*}{\textbf{Sub-approach}}} & \multicolumn{1}{l}{\textbf{Item removal}} & \textbf{Feature removal} & \textbf{Class removal} & \textbf{Task removal} & \textbf{Stream removal} & \multicolumn{1}{l}{\textbf{Recoverable}}     & \textbf{Irrecoverable}   & \textbf{Exact}           & \multicolumn{1}{l}{\textbf{Approximate}}     & \multicolumn{1}{c}{\multirow{-2}{*}{\textbf{Related work}}}                                                                                   \\
\midrule
                                                               &                                                           &                                                             & Feature   Statistics-Based                                  & \multicolumn{1}{l}{}                      & \multicolumn{1}{c}{$\checkmark$}    &                        &                       &                         & \multicolumn{1}{l}{NA} & NA & NA & \multicolumn{1}{l}{NA} & \cite{Gretton2012,Bernico2019,Sun2021,Zhang2022,Lin2017,Ying2018,Gong2012,Zhang2020}                                                          \\
                                                               &                                                           &                                                             & Test   Prediction-Based                                     & \multicolumn{1}{l}{}                      &                          &                        & \multicolumn{1}{c}{$\checkmark$} &                         & \multicolumn{1}{l}{NA} & NA & NA & \multicolumn{1}{l}{NA} & \cite{Xie2017,Sun2018a,Ben-David2010,Xu2018,Wu2020c,Feng2021}                                                                                 \\
                                                               &                                                           & \multirow{-3}{*}{Domain   Similarity Estimation Approaches} & Fine-Tuning-Based                                           & $\checkmark$                                         & \multicolumn{1}{c}{$\checkmark$}    & \multicolumn{1}{c}{$\checkmark$}  &                       &                         & \multicolumn{1}{l}{NA} & NA & NA & \multicolumn{1}{l}{NA} & \cite{Tran2019,Nguyen2020a,Afridi2018,Huang2022,Bao2019,Madasu2021}                                                                           \\
                                                               \clineB{3-4}{0.2}
                                                               &                                                           &                                                             & Data Transferability Enhancement                            & $\checkmark$                                         & \multicolumn{1}{c}{$\checkmark$}    & \multicolumn{1}{c}{$\checkmark$}  &                       &                         & $\checkmark$                                            & \multicolumn{1}{c}{$\checkmark$}    & \multicolumn{1}{c}{$\checkmark$}    & \multicolumn{1}{l}{}                         & \cite{Ahmed2021,Turrisi2022,Kim2020,Zhu2020,Seah2012,Wu2018,Xu2020,Yang2022,DeMathelin2021,Li2022a,Chen2019b,Yosinski2014,Lipton2018,You2019} \\
                                                               &                                                           &                                                             & Model   Transferability Enhancement                         & \multicolumn{1}{l}{}                      &                          &                        & \multicolumn{1}{c}{$\checkmark$} &                         & \multicolumn{1}{l}{}                         & \multicolumn{1}{c}{$\checkmark$}    &                          & $\checkmark$                                            & \cite{Madry2017,Zhang2021,Liu2021,Han2022,You2020,Ishii2021,Gao2021}                                                                          \\
                                                               &                                                           &                                                             & Training   Process Enhancement                              & \multicolumn{1}{l}{}                      &                          &                        & \multicolumn{1}{c}{$\checkmark$} & \multicolumn{1}{c}{$\checkmark$}   & $\checkmark$                                            &                          &                          & $\checkmark$                                            & \cite{Long2013,Long2015,Yoon2018,Xiao2021,Saito2021,Liang2021,Wan2019,Yoon2018,Wan2019,Xiao2021,Liang2021}                                    \\
                                                               & \multirow{-7}{*}{\textbf{Adaptability}}                   & \multirow{-4}{*}{NT mitigation approaches}                  & Target   Prediction Enhancement                             & \multicolumn{1}{l}{}                      & \multicolumn{1}{c}{$\checkmark$}    & \multicolumn{1}{c}{$\checkmark$}  & \multicolumn{1}{c}{$\checkmark$} &                         & \multicolumn{1}{l}{}                         & \multicolumn{1}{c}{$\checkmark$}    &                          & $\checkmark$                                            & \cite{Pei2018,Gui2018,Wang2020a,Liang2020,Tian2020,Jin2020}                                                                                   \\
                                                               \cline{2-14}
                                                               &                                                           &                                                             & Knowledge Evolution (KE)                                    & $\checkmark$                                         & \multicolumn{1}{c}{$\checkmark$}    &                        & \multicolumn{1}{c}{$\checkmark$} &                         & $\checkmark$                                            &                          &                          & $\checkmark$                                            & \cite{Taha2021}                                                                                                                               \\
                                                               &                                                           &                                                             & Iterative   Magnitude Pruning (IMP)                         & $\checkmark$                                         & \multicolumn{1}{c}{$\checkmark$}    &                        & \multicolumn{1}{c}{$\checkmark$} &                         & $\checkmark$                                            &                          &                          & $\checkmark$                                            & \cite{Frankle2018,Frankle2019}                                                                                                                \\
                                                               &                                                           &                                                             & Reinitializing   Final Layer (RIFLE)                        & $\checkmark$                                         & \multicolumn{1}{c}{$\checkmark$}    &                        & \multicolumn{1}{c}{$\checkmark$} &                         & $\checkmark$                                            &                          &                          & $\checkmark$                                            & \cite{Li2020b}                                                                                                                                \\
                                                               &                                                           &                                                             & Later-Layer   Forgetting (LLF)                              & $\checkmark$                                         & \multicolumn{1}{c}{$\checkmark$}    &                        & \multicolumn{1}{c}{$\checkmark$} &                         & \multicolumn{1}{l}{}                         & \multicolumn{1}{c}{$\checkmark$}    & \multicolumn{1}{c}{$\checkmark$}    & \multicolumn{1}{l}{}                         & \cite{Zhou2022}                                                                                                                               \\
                                                               &                                                           &                                                             & Iterated   Learning                                         & $\checkmark$                                         & \multicolumn{1}{c}{$\checkmark$}    & \multicolumn{1}{c}{$\checkmark$}  & \multicolumn{1}{c}{$\checkmark$} &                         & $\checkmark$                                            &                          &                          & $\checkmark$                                            & \cite{Kirby2001,Kirby2014,Hoang2018,Ren2020a}                                                                                                 \\
                                                               & \multirow{-6}{*}{\textbf{Generalisation}}                 & \multirow{-6}{*}{Iterative   training}                      & Partial   Balanced Forgetting (PBF)                         & $\checkmark$                                         & \multicolumn{1}{c}{$\checkmark$}    &                        & \multicolumn{1}{c}{$\checkmark$} &                         & \multicolumn{1}{l}{}                         & \multicolumn{1}{c}{$\checkmark$}    & \multicolumn{1}{c}{$\checkmark$}    & \multicolumn{1}{l}{}                         & \cite{Li2019a,Zhou2022}                                                                                                                       \\
                                                               \cline{2-14}
                                                               &                                                           &                                                             & Dynamic attention spans                                     & \multicolumn{1}{l}{}                      & \multicolumn{1}{c}{$\checkmark$}    &                        & \multicolumn{1}{c}{$\checkmark$} &                         & $\checkmark$                                            &                          &                          & $\checkmark$                                            & \cite{Sukhbaatar2019,Correia2019,   Wu2022b}                                                                                                  \\
                                                               &                                                           & \multirow{-2}{*}{Improve attention's   efficiency}          & Attention   with Bounded-Memory (ABC)                       & \multicolumn{1}{l}{}                      & \multicolumn{1}{c}{$\checkmark$}    &                        &                       &                         & $\checkmark$                                            & \multicolumn{1}{c}{$\checkmark$}    &                          & $\checkmark$                                            & \cite{Peng2021,Peng2021a}                                                                                                                     \\
                                                               \clineB{3-4}{0.2}
                                                               &                                                           &                                                             & Expire-span                                                 & $\checkmark$                                         & \multicolumn{1}{c}{$\checkmark$}    &                        & \multicolumn{1}{c}{$\checkmark$} &                         & \multicolumn{1}{l}{}                         & \multicolumn{1}{c}{$\checkmark$}    & \multicolumn{1}{c}{$\checkmark$}    & \multicolumn{1}{l}{}                         & \cite{Sukhbaatar2021}                                                                                                                         \\
                                                               &                                                           &                                                             & Compressor   (RECOMP, TRIME)                                & $\checkmark$                                         & \multicolumn{1}{c}{$\checkmark$}    &                        & \multicolumn{1}{c}{$\checkmark$} &                         & $\checkmark$                                            & \multicolumn{1}{c}{$\checkmark$}    & \multicolumn{1}{c}{$\checkmark$}    & $\checkmark$                                            & \cite{Xu2023,Zhong2022}                                                                                                                       \\
\multirow{-18}{*}{\textbf{Active Forgetting}}                  & \multirow{-5}{*}{\textbf{Compression}}                    & \multirow{-3}{*}{Lossless compression}                      & Neural   Attentive Filtering (AI2V++)                       & $\checkmark$                                         & \multicolumn{1}{c}{$\checkmark$}    &                        & \multicolumn{1}{c}{$\checkmark$} &                         & $\checkmark$                                            &                          &                          & $\checkmark$                                            & \cite{Gaiger2023}                                                                                                                             \\
\cline{1-14}
                                                               &                                                           & Statistical Query Learning                                  & SQ learning                                                 & $\checkmark$                                         &                          &                        &                       &                         & \multicolumn{1}{l}{}                         &                          & \multicolumn{1}{c}{$\checkmark$}    & \multicolumn{1}{l}{}                         & \cite{Cao2015}                                                                                                                                \\
                                                               \clineB{3-4}{0.2}
                                                               &                                                           &                                                             & SISA                                                        & $\checkmark$                                         &                          &                        &                       &                         & $\checkmark$                                            &                          & \multicolumn{1}{c}{$\checkmark$}    & \multicolumn{1}{l}{}                         & \cite{Bourtoule2021,Aldaghri2021,He2021}                                                                                                      \\
                                                               &                                                           &                                                             & Graph Partition                                             & $\checkmark$                                         &                          &                        &                       &                         & $\checkmark$                                            &                          & \multicolumn{1}{c}{$\checkmark$}    & \multicolumn{1}{l}{}                         & \cite{Chen2022a}                                                                                                                              \\
                                                               &                                                           & \multirow{-3}{*}{Data Partitioning}                         & Data partition in Recommender systems                       & $\checkmark$                                         &                          &                        &                       &                         & $\checkmark$                                            &                          & \multicolumn{1}{c}{$\checkmark$}    & \multicolumn{1}{l}{}                         & \cite{Chen2022}                                                                                                                               \\
                                                               \clineB{3-4}{0.2}
                                                               &                                                           &                                                             & Deltagrad                                                   & \multicolumn{1}{l}{}                      &                          &                        & \multicolumn{1}{c}{$\checkmark$} &                         & $\checkmark$                                            &                          & \multicolumn{1}{c}{$\checkmark$}    & \multicolumn{1}{l}{}                         & \cite{Wu2020}                                                                                                                                 \\
                                                               &                                                           & \multirow{-2}{*}{Rapid Retraining}                          & ARCANE                                                      & \multicolumn{1}{l}{}                      &                          &                        & \multicolumn{1}{c}{$\checkmark$} &                         & $\checkmark$                                            &                          & \multicolumn{1}{c}{$\checkmark$}    & \multicolumn{1}{l}{}                         & \cite{Yan2022}                                                                                                                                \\
                                                               \clineB{3-4}{0.2}
                                                               &                                                           &                                                             & Certified removal                                           & $\checkmark$                                         &                          &                        &                       & \multicolumn{1}{c}{$\checkmark$}   & \multicolumn{1}{l}{}                         &                          & \multicolumn{1}{c}{$\checkmark$}    & \multicolumn{1}{l}{}                         & \cite{Ullah2021}                                                                                                                              \\
                                                               &                                                           &                                                             & Generalisation theory                                       & \multicolumn{1}{l}{}                      &                          &                        &                       &                         & \multicolumn{1}{l}{}                         &                          & \multicolumn{1}{c}{$\checkmark$}    & \multicolumn{1}{l}{}                         & \cite{Sekhari2021}                                                                                                                            \\
                                                               &                                                           &                                                             & Bayesian Inferrence                                         & $\checkmark$                                         &                          &                        &                       &                         & \multicolumn{1}{l}{}                         &                          & \multicolumn{1}{c}{$\checkmark$}    & \multicolumn{1}{l}{}                         & \cite{Fu2021,Nguyen2022a}                                                                                                                     \\
                                                               & \multirow{-10}{*}{\textbf{Exact   machine unlearning}}    & \multirow{-4}{*}{Others}                                    & Linear GNN                                                  & \multicolumn{1}{l}{}                      &                          &                        &                       &                         & \multicolumn{1}{l}{}                         &                          & \multicolumn{1}{c}{$\checkmark$}    & \multicolumn{1}{l}{}                         & \cite{Cong2023}                                                                                                                               \\
                                                               \cline{2-14}
                                                               &                                                           &                                                             & Differencial Privacy                                        & $\checkmark$                                         & \multicolumn{1}{c}{$\checkmark$}    &                        &                       & \multicolumn{1}{c}{$\checkmark$}   & $\checkmark$                                            & \multicolumn{1}{c}{}     &                          & $\checkmark$                                            & \cite{Golatkar2020,Golatkar2020a}                                                                                                             \\
                                                               &                                                           &                                                             & Mixed-privacy setting                                       & $\checkmark$                                         &                          &                        &                       &                         & $\checkmark$                                            & \multicolumn{1}{c}{}     &                          & $\checkmark$                                            & \cite{Golatkar2021}                                                                                                                           \\
                                                               &                                                           & \multirow{-3}{*}{Weight Scrubbing}                          & Fisher Information Matrix                                   & \multicolumn{1}{l}{}                      &                          &                        &                       &                         & $\checkmark$                                            & \multicolumn{1}{c}{}     &                          & $\checkmark$                                            & \cite{Martens2020}                                                                                                                            \\
                                                               \clineB{3-4}{0.2}
                                                               &                                                           &                                                             & Influence method                                            & $\checkmark$                                         &                          &                        &                       &                         & $\checkmark$                                            & \multicolumn{1}{c}{}     &                          & $\checkmark$                                            & \cite{Guo2019,Izzo2021,Chien2022,Wu2022,Warnecke2021}                                                                                         \\
                                                               &                                                           &                                                             & Adding regularization terms                                 & \multicolumn{1}{l}{}                      &                          &                        &                       &                         & $\checkmark$                                            & \multicolumn{1}{c}{}     &                          & $\checkmark$                                            & \cite{Zeng2021LearningTR}                                                                                                                     \\
                                                               &                                                           & \multirow{-3}{*}{Data Influence}                            & Fisher-based   unlearning method                            & \multicolumn{1}{l}{}                      &                          &                        &                       &                         & $\checkmark$                                            & \multicolumn{1}{c}{}     &                          & $\checkmark$                                            & \cite{Peste2021}                                                                                                                              \\
                                                               \clineB{3-4}{0.2}
                                                               &                                                           &                                                             & Gradient based method                                       & $\checkmark$                                         & \multicolumn{1}{c}{}     &                        &                       & \multicolumn{1}{c}{$\checkmark$}   & $\checkmark$                                            & \multicolumn{1}{c}{}     &                          & $\checkmark$                                            & \cite{Neel2021}                                                                                                                               \\
                                                               &                                                           &                                                             & Incremental learning                                        & $\checkmark$                                         &                          &                        &                       &                         & $\checkmark$                                            & \multicolumn{1}{c}{}     &                          & $\checkmark$                                            & \cite{Wu2020a}                                                                                                                                \\
\multirow{-19}{*}{\textbf{Passive Forgetting}}                 & \multirow{-9}{*}{\textbf{Approximate machine unlearning}} & \multirow{-3}{*}{Updates Control}                           & Amnesiac unlearning                                         & \multicolumn{1}{l}{}                      & \multicolumn{1}{c}{$\checkmark$}    & \multicolumn{1}{c}{$\checkmark$}  & \multicolumn{1}{c}{$\checkmark$} &                         & $\checkmark$                                            & \multicolumn{1}{c}{}     &                          & $\checkmark$                                            & \cite{Graves2021}\\
\bottomrule
\end{tabular}
}
\caption{Summary of forgetting approaches in Machine Learning (NA means the field is not applicable)}
\label{tab:forgetting_landscape}
\end{sidewaystable}

\subsection{Active forgetting approaches}
According to their purposes, existing studies about active forgetting could be categorised as forgetting to mitigate negative transfer; forgetting to enhance the generalisation ability to unseen data; and forgetting to free up storage space thus enabling the transformer to scale. This survey will expand the discussion about active forgetting approaches in machine learning according to the above distinct scenarios. The structure of this chapter with some corresponding relevant work is presented in Table \ref{tab:forgetting_landscape}.

\subsubsection{Domain similarity estimation}
One major reason for NT to happen is substantial domain divergence \cite{Zhang2023a}, which refers to significant differences in data distributions, feature spaces, and inherent characteristics between the source domain (where the model is initially trained) and the target domain (where the model is subsequently applied or fine-tuned). Therefore, before implementing an active forgetting strategy to mitigate negative transfer learning, it's essential to first estimate domain similarity. As summarised by \cite{Zhang2023a}, existing estimation methods encompass feature statistics-based approaches, test prediction-based methods and fine-tuning-based techniques.

\textbf{Feature statistics-based approaches} in transfer learning analyse data distributions and statistics (mean, variance) in source and target domains to estimate domain similarity. Techniques include histogram comparisons, correlation coefficients, and covariance matrix analyses to measure distribution discrepancies, with popular methods includes Earth Mover's Distance, Maximum Mean Discrepancy, and KL-divergence.

\textbf{Test-prediction-based approaches} evaluate model performance from source to target domain, utilising strategies like target performance, domain classifier, and consensus focus. These assess domain similarity through model accuracy on labelled target data, classification error from a domain classifier, and agreement among various source models on target predictions. 

\textbf{Fine-tuning-based approaches} adjust a pre-trained model to a new target domain by modifying higher layer parameters while retaining lower layer weights. Methods include label-based estimation, assessing domain similarity by comparing labels, and combined label-feature representation estimation, which also considers feature alignment post-fine-tuning. Techniques to prevent catastrophic forgetting, like L1/L2 regularisation and Elastic Weight Consolidation, optimise knowledge transfer by balancing source knowledge retention and target domain adaptation. Fine-tuning-based approaches might involve various aspects like 'Item removal', 'Feature removal', and 'Class removal', depending on the specifics of the fine-tuning process.

\subsubsection{NT mitigation approaches}
Negative Transfer (NT) mitigation focuses on minimising the detrimental effects that occur when inappropriate knowledge is transferred between domains. This process is effectively managed through active forgetting, where specific knowledge is intentionally discarded or de-emphasised to enhance outcomes in the target domain \cite{Zhang2023a}.

To address \textbf{data transferability}, several strategies across different dimensions are utilised to mitigate NT. The domain dimension involves strategies like domain selection and weighting via convex optimisation, which determines the relevance of different domains to ensure the transfer of pertinent information \cite{Ahmed2021,Turrisi2022}. Additionally, domain decomposition helps by strategically segmenting a domain into subdomains to facilitate more focused adaptation \cite{Kim2020,Zhu2020}. Moving to the instance dimension, techniques such as Predictive Distribution Matching (PDM) and active learning prioritise instances from the source domain that are beneficial for the target domain, implicitly omitting less salient instances \cite{Seah2012,Wu2018}. This selection process helps to "forget" non-representative data, streamlining the knowledge transfer. The feature dimension introduces complexity by breaking down or adjusting the feature space to emphasise features that are universally applicable. Methods like Dual TL and dynamic auxiliary importance classifiers are employed to highlight and retain beneficial features while sidelining those that do not contribute positively \cite{DeMathelin2021,Li2022a}. In the class dimension, shift detection and weighted adversarial networks ensure that the classes align between the source and target domains. Techniques like universal domain adaptation actively filter out unrelated source data that does not conform to the target domain's class structure, effectively discarding non-aligning data \cite{Lipton2018, You2019}.

For \textbf{model transferability}, enhancing source model training through methods such as transferable batch normalization allows the model to prioritise more adaptable features and "forget" less relevant ones \cite{Wang2019d}. Target model adaptation then focuses on selecting transferable parameters and applying parameter regularization to further reduce the influence of inappropriate data or features \cite{Han2022}.

\textbf{Training process enhancement} involves managing training dynamics, such as varying step sizes and directions during optimization. This is often guided by a hyper-parameter, $\lambda$, which helps balance between discriminability and transferability \cite{Yoon2018, Xiao2021}.

\textbf{Target prediction enhancement} employs techniques like soft pseudo-labelling, selective pseudo-labelling, clustering-enhanced pseudo-labelling, and entropy regularization. These methods enable the model to adapt to new domains by "forgetting" noisy or irrelevant labels and data, thus refining the prediction accuracy in the target domain \cite{Pei2018, Wang2020a, Gui2018, Tian2020, Jin2020, Grandvalet2004, Liang2020}.

\subsubsection{Iterative training}
Recent training algorithms have highlighted the importance of iteratively refining machine learning models to enhance their generalization ability \cite{Frankle2018,Frankle2019,Zhou2022}. Techniques such as "Knowledge Evolution" involve reinitializing parts of the network while consistently training others to refine the learning process \cite{Taha2021}. Similarly, "Iterative Magnitude Pruning" involves a cycle of pruning and retraining, which has shown to improve performance by selectively removing weights \cite{Frankle2018,Frankle2019}.

Li et al. introduced RIFLE, a method that periodically reinitializes the final layer of the model to reduce dependency on specific features, thereby encouraging the learning of broader patterns \cite{Li2020b}. This approach, along with strategies utilizing synthetic machine translation corpora and iterative self-distillation, effectively enhances learning by leveraging back-translations and successive teaching iterations \cite{Hoang2018,Furlanello2018,Xie2020a}.

Particularly noteworthy is the "Iterated Learning" approach, which has been effective in enhancing compositionality and mitigating language drift in emergent languages, as demonstrated in various studies \cite{Kirby2001,Ren2020a,Vani2021,Lu2020}. Zhou et al. proposed a "later-layer forgetting" (LLF) strategy, which involves reinitializing later layers of the network to specifically remove information associated with complex examples, thereby promoting a "forget-and-relearn" paradigm. This method encourages relearning difficult examples using simpler features from early layers, which has proven to enhance generalization in image classification and emergent communication tasks \cite{Zhou2022}.

\begin{figure}[h]
\centering
\begin{minipage}{0.4\textwidth}
\centering
\includegraphics[width=\linewidth]{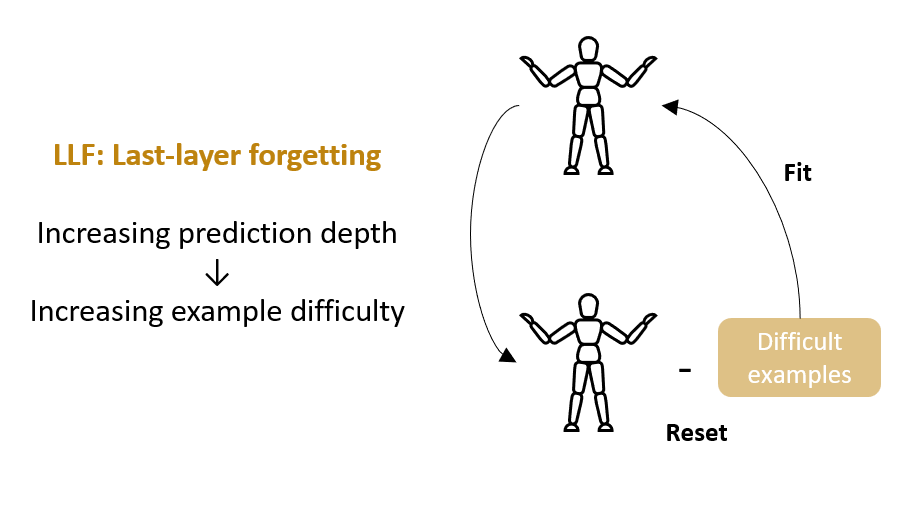}
\caption{Last-layer forgetting (LLF)}
\label{fig
}
\end{minipage}%
\begin{minipage}{0.4\textwidth}
\centering
\includegraphics[width=\linewidth]{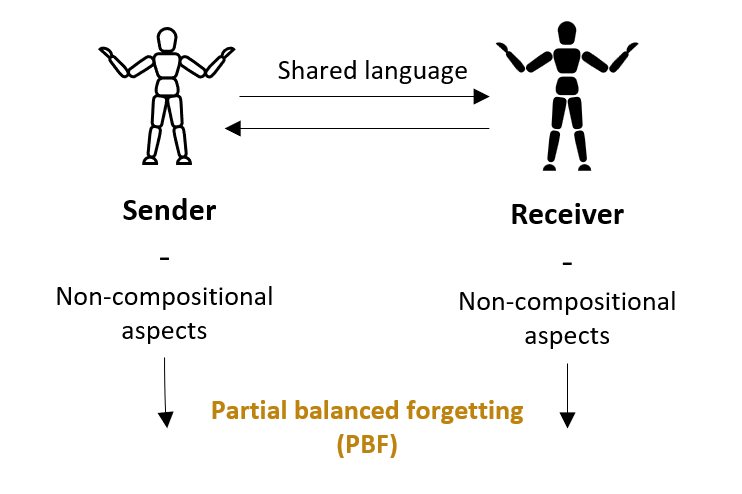}
\caption{Partial balanced forgetting (PBF)}
\label{fig
}
\end{minipage}
\end{figure}

The iterative retraining process is vital for refining emergent languages, where consistently understood features across training "generations" become reinforced. Unlike traditional training phases that focus on imitation, iterated learning thrives on retraining interactions, promoting structure-preserving mappings as seen in human linguistics \cite{Kirby2001,Kirby2014}.

In emergent language research, the focus spans foundational concepts to the dynamics of sender-receiver interactions. Central to this is selective forgetting, which enhances generalization in language learning models. By periodically reinitializing receivers, the model promotes compositionality, as a fresh receiver is more likely to adapt to a structured language, leading to improved communication efficiency \cite{Li2019a,Zhou2022}.

This approach aligns with the concept of "Partial Balanced Forgetting" (PBF), where both sender and receiver selectively forget non-compositional aspects of the language to enhance overall structure and generalization. This strategic forgetting helps tackle complex language tasks by pushing the model towards summarizing and abstracting information, rather than getting overwhelmed by complexities \cite{Ren2020a}.

\subsubsection{Improve attention's efficiency}
Recent research has focused on optimising attention mechanisms to facilitate more concentrated learning. \cite{Bhattacharjee2022} proposed a mathematical approach that uses a model where a learner decides which facts to store, and the system measures performance based on how it matches up to various experts with specific memory preferences. Dynamic attention spans, such as Adaptive Span \cite{Sukhbaatar2019} and Adaptive Sparse \cite{Correia2019} focus on learning which attention heads can have shorter spans. Instead of attending to all tokens, the model dynamically chooses which tokens to attend to, thereby reducing the computational overhead. This selective attention mechanism can be seen as a form of "forgetting," where the model disregards less relevant tokens in favour of more meaningful ones. Other research focus on attenuating attention through employing fixed attention masks, sliding windows, and hierarchical architectures. \cite{Wu2022b} offers an innovative approach to scale transformers by introducing kNN lookup and non-differentiable external memory. These techniques allow the model to efficiently manage its attention mechanism, focusing on essential details while "forgetting" unnecessary computations. 
The various control strategies that are used to organise consolidated memories are summarised as a unified abstraction: attention with bounded-memory control (ABC) \cite{Peng2021}. \(ABC_{MLP}\) introduced in \cite{Peng2021a} as an instance of ABC introduces a neural network to determine how to store each token in the memory to strike a better trade-off between accuracy and efficiency.
\subsubsection{Lossless compression}
\begin{figure}
    \centering
    \includegraphics[width=0.5\linewidth]{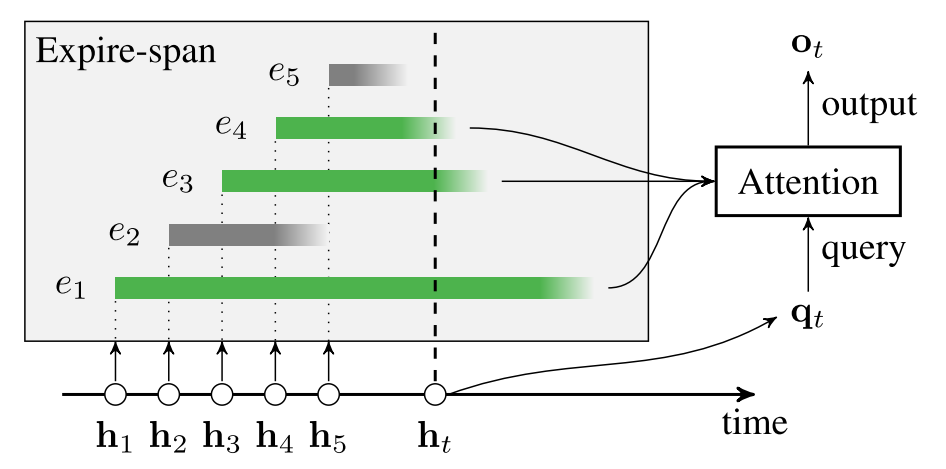}
    \caption{Expire-Span \cite{Sukhbaatar2021}. For every memory \( h_i \), we compute an \textsc{expire-span} \( e_i \) that determines how long it should stay in memory. Here, memories \( h_2 \) and \( h_5 \) are already expired at time \( t \), so the query \( q_t \) can only access \(\{ h_1, h_3, h_4 \}\) in self-attention.}
    \label{fig:expire-span}
\end{figure}

The aforementioned methods primarily aim to make attention more efficient without minimizing memory usage. To address this, Expire-span \cite{Sukhbaatar2021} was introduced, calculating a specific lifespan for each memory (Figure \ref{fig:expire-span}). Once a memory surpasses its lifespan, it becomes inaccessible. This strategy prunes distant tokens deemed "unimportant", allowing transformers to handle memories tens of thousands in size. This temporal forgetting means transformers retain only recent and pertinent data, thereby cutting memory demands and computational costs.
The principle of lossless compression resonates with the transformer's goal of maximizing information retention with minimal resources. Compression techniques, similar to the act of forgetting, aspire to condense data without significant information loss. In transformers, the attention mechanism, optimised via methods like Adaptive-Span or Expire-span, is akin to data compression, retaining only vital information. 

The principle behind the compressor serves various applications, with its primary use in language model training. RECOMP \cite{Xu2023} is proposed for optical language model performance and operates by first retrieving and then compressing documents into summaries before assimilation. The model uses two compressor types: an extractive one that selects key sentences and an abstractive one that generates summaries from multiple sources. These compressors, designed to enhance the language model's task performance, can also opt for "selective augmentation" by returning an empty string if the documents are irrelevant. This selective approach mirrors the "forgetting" concept in transformers, streamlining information input to ensure optimal trade-off between model accuracy and efficiency. Meanwhile, TRIME \cite{Zhong2022} emphasizes memory type classification, introducing methods for memory construction and data batching for different memory types during testing. The attentive Item2Vec++ (AI2V++) model proposed a neural attentive collaborative filtering approach in which the user representation adapts dynamically in the presence of the recommended item \cite{Gaiger2023}. Inspired by the dynamic nature of the human memory, in which different memories become more accessible in different contexts, the model uses attention mechanisms to dynamically adjust the importance of the historical items based on the item being considered for recommendation. 

\subsection{Passive forgetting approaches}
In this survey, unlearning is categorised as a subset of machine forgetting because it only includes a passive forgetting process triggered by removal requests due to privacy, security or usability concerns. Previous research in machine unlearning has consistently classified the methods into "Exact (complete, perfect) unlearning" and "Approximate (certified, bounded) unlearning", which aligns with our prior definitions of forgetting dimensions \ref{subsection: Extent of forgetting}.

\subsubsection{Exact machine unlearning}
The most straightforward way to perform exact machine unlearning is by removing the requested "forget set" and retraining the model from scratch, which is computationally demanding. Recent studies have explored various exact unlearning strategies to make the retraining process both efficient and easy to control. \textbf{Statistical Query Learning, Data Partitioning, and Rapid Retraining} appear to be the most related approaches for exact unlearning \cite{Nguyen2022,Zhang2023,Cong2023,Wang2023}. Therefore, we will introduce those methods in more detail and cover other approaches more briefly (please refer to \cite{Nguyen2022} for a more in-depth summarisation of unlearning methodologies.

\paragraph{Statistical Query Learning}
Statistical Query (SQ) learning trains models using statistical data properties instead of direct data access \cite{Kearns1998}. This method supports efficient unlearning by updating statistical summaries without revisiting individual data points. Cao and Yang \cite{Cao2015} applied SQ learning for machine unlearning by updating summations, reducing data dependencies. However, this approach struggles with complex models like deep neural networks due to the exponential increase in the required statistical queries \cite{Bourtoule2021}, which impacts the efficiency of unlearning and relearning processes.

\begin{figure}
    \centering
    \includegraphics[width=0.5\linewidth]{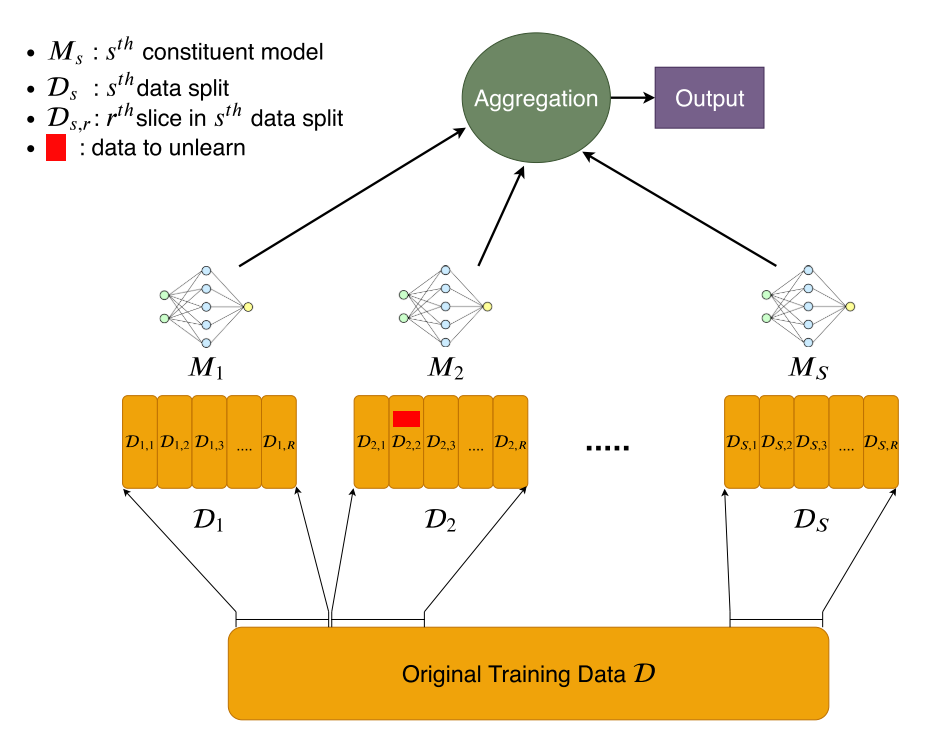}
    \caption{SISA proposed by \cite{Bourtoule2021}. It splits the dataset into multiple shards and train an independent model on each data shard, then aggregate their prediction during inference.}
    \label{fig:SISA}
\end{figure}

\paragraph{Data Partition}
To reduce computational costs, Bourtoule et al. \cite{Bourtoule2021} introduced the SISA training approach, which divides the dataset into multiple shards, each trained independently and then aggregated during inference (see Figure \ref{fig:SISA}). This method allows for targeted re-training upon unlearning requests, enhancing process efficiency. However, model performance may decline due to reduced training data per shard and data heterogeneity. Complex learning tasks may require integrating SISA with other methods like transfer learning to maintain accuracy and processing speed.

\paragraph{Rapid Retraining}
Innovations in exact unlearning focus on accelerating the retraining process. Deltagrad \cite{Wu2020} caches model parameters and gradients to expedite retraining after data deletion. ARCANE \cite{Yan2022} uses ensemble learning to transform full retraining into simpler classification tasks, reducing computational demands while maintaining performance. Additionally, Ullah et al. \cite{Ullah2021} and others explore efficient retraining strategies and theoretical unlearning approaches for various applications including Bayesian inference \cite{Fu2021, Nguyen2022a} and algorithmic unlearning \cite{Cong2023}.

\subsubsection{Approximate machine unlearning}
\label{Approximate Unlearning}
Approximate unlearning has been proposed to mitigate the computationally intensive nature of exact unlearning. The central concept involves approximating the model trained without the deleted data in the parameter space. Mahadevan et al. \cite{Mahadevan2021} and Zhang et al. \cite{Zhang2023} summarise that approximate unlearning methods can be broadly categorised into three groups. This subsection will introduce several of the most relevant methods based on this classification.

\paragraph{Weight 'Scrubbing'}
Differential privacy, based on the Fisher Information Matrix (FIM), is used in approximate unlearning strategies to maintain certifiability when updating machine learning models. This method, known as Fisher forgetting, involves adjusting the weights to obscure specific training data influences, avoiding the need for complete retraining \cite{Martens2020,Golatkar2020a, Golatkar2020, Golatkar2021}. Golatkar et al. extend this to a mixed-privacy setting, allowing some training samples to remain unaltered, simplifying the unlearning process for non-core data by setting certain weights to zero, which minimally impacts performance.

\paragraph{Data Influence}
Unlearning approaches often utilize influence functions to understand how changes in training data affect model parameters. These functions help compute the impact of deleted data to ensure it does not influence the model after its removal. Techniques range from employing first-order Taylor approximations to using influence functions for estimating training data impacts on parameters, enabling updates without reprocessing the entire dataset \cite{Chundawat2023, Mahadevan2022, Guo2019, Izzo2021, Chien2022, Wu2022, Warnecke2021}. However, ensuring that data meant to be forgotten remains entirely unrepresented in the model remains a challenge \cite{Thudi2022a}.
\paragraph{Updates control}
Influencing model training via updates control involves tracking and potentially reversing training steps to remove data influences. This method, known as Amnesiac Unlearning, selectively undoes updates involving sensitive data. Techniques involve saving intermediate training states like weight parameters and gradients to efficiently recalibrate the model’s training path upon data deletion requests. While effective, this approach does not guarantee perfect data removal and often involves adding random noise to achieve approximate unlearning \cite{Graves2021, Neel2021, Wu2020, Wu2020a, Thudi2022}.

\section{Discussion}
\label{sec6: Discussion}
\subsection{Applications of Forgetting Approaches in ML}

Active forgetting in machine learning involves selectively ignoring outdated or misleading knowledge to enhance model performance across various fields. In the medical and healthcare sector, this approach improves diagnostic accuracy by disregarding non-relevant features from different demographics or imaging devices, thus enhancing model transferability \cite{Niu2021,Zhang2020,PeresdaSilva2021,He2019}. In sentiment analysis, the mechanism prevents misinterpretation of sentiments by ignoring domain-specific jargon that is unsuitable for different contexts \cite{Liu2019}. For autonomous driving, it adapts vehicles to different driving conditions by forgetting irrelevant driving patterns learned in other regions \cite{Sorocky2020,Corso2021}. Similarly, in robotics, active forgetting enhances robot adaptability to new tasks by discarding behaviors specific to previous environments \cite{Li2020a}.

Passive forgetting, or machine unlearning, is critical for addressing security, privacy, and biases in machine learning by removing specific data influences from models. In healthcare, this ensures the accuracy and security of models handling sensitive data, supports privacy compliance, and facilitates federated learning without needing to share patient data directly \cite{Nguyen2022,Ren2020,Marchant2022,Wu2022a}. In cybersecurity, machine unlearning counteracts backdoor and membership inference attacks, enhancing model integrity and trustworthiness \cite{Toreini2020,Wang2019a,Liu2022,Chen2021}. Furthermore, machine unlearning plays a significant role in correcting social biases by removing biased data, thereby promoting fairer outcomes in sensitive applications such as hiring and law enforcement \cite{Mehrabi2021,Zou2018,Osoba2017,Kumar2022}.

\subsection{Challenges}
\subsubsection{Model Training}
Selecting data to forget, known as \textbf{"forget set" selection}, is a pivotal challenge in machine learning, particularly in active forgetting scenarios. This process involves evaluating the utility of each data point to determine their impact on the model's parameters. Core-set selection simplifies this by identifying a subset that yields comparable results to the full dataset, thus reducing learning costs and enhancing efficiency in forgetting \cite{Huang2010,Sener2017,Baykal2018}. The main difficulty lies in accurately defining this core-set to capture all essential information without compromising model performance \cite{Bourtoule2021}. Additionally, deciding the granularity of forgetting—whether whole categories or specific instances—requires understanding data dependencies to avoid residual distortions in model output.

\textbf{Training models to forget} involves significant complexity due to the unpredictable nature of neural network training, which uses random mini-batches and sequences. This randomness disperses the influence of individual data points, making it hard to isolate their specific impacts. Forgetting requires not just removing the data point but also negating its cumulative effects to maintain consistent model performance. This process is computationally intensive, particularly in exact unlearning scenarios where parts of the model may need frequent retraining. These challenges are exacerbated in resource-limited environments, such as in online continual learning and meta-learning, where the retention of previously learned information is further complicated by limited data exposure \cite{Bourtoule2021, Nguyen2022, Wang2023}.

\textbf{Catastrophic forgetting} is a significant challenge in continual learning. As new data is introduced, the model's ability to perform tasks related to older data can substantially degrade. This problem is compounded in settings with continual and federated learning due to potential task interference. Various strategies, such as designing special loss functions, have been proposed to mitigate catastrophic forgetting, though challenges remain in deciding which data points to replicate for improving model accuracy without increasing the overhead of forgetting \cite{Nguyen2020, Du2019, Golatkar2020, Settles2009}.

\subsubsection{Evaluation and monitoring}
\label{Evaluation and monitoring}
Evaluating machine forgetting involves \textbf{key metrics} such as accuracy, completeness, and efficiency. Accuracy assesses how well the model performs post-forgetting, aiming to match the precision of a freshly retrained model on various datasets \cite{Mahadevan2022, Nguyen2022}. Completeness ensures that the forgotten data leaves no residual trace, aligning the predictions of the unlearned model with those of a model trained from scratch. This is crucial for privacy, as it ensures that data once forgotten remains inaccessible, even with sophisticated Membership Inference Attacks (MIAs) or model inversion techniques \cite{Carlini2022, Graves2021}. Efficiency measures the resource consumption involved in the forgetting process, considering factors like time and computational overhead. Efficient unlearning is essential to manage costs and maintain the feasibility of forgetting processes, especially in resource-limited settings \cite{Izzo2021, Nguyen2022}. Equally important is ensuring that the evaluation metrics themselves are computed efficiently, not just the forgetting algorithms. As this field is still emerging, there are many relevant facets that could be incorporated into the forgetting evaluation metric, yet they have not been explored by current research. These include the applicability of the forgetting method to different models, the granularity of forgetting depths, the rate at which forgetting occurs, and the time decay that accounts for the fading of information over time, especially in models set for continual learning. A more comprehensive, efficient, and reliable evaluation metric for machine forgetting is needed.

\textbf{Data lineage management} tracks the flow and transformation of data throughout its lifecycle in a system, from origin to end-use. It is vital for ensuring accurate and compliant machine forgetting, especially under regulations like GDPR which mandate the right to be forgotten. This process enables precise identification and removal of data, fostering trust among users by transparently showing that their data can be completely erased upon request. Additionally, data lineage is crucial for troubleshooting and reproducibility, helping to pinpoint issues or reproduce outcomes when models are retrained without certain datasets \cite{Li2022, Zhang2023, Baracaldo2017}.
\begin{figure}
    \centering
    \includegraphics[width=0.75\linewidth]{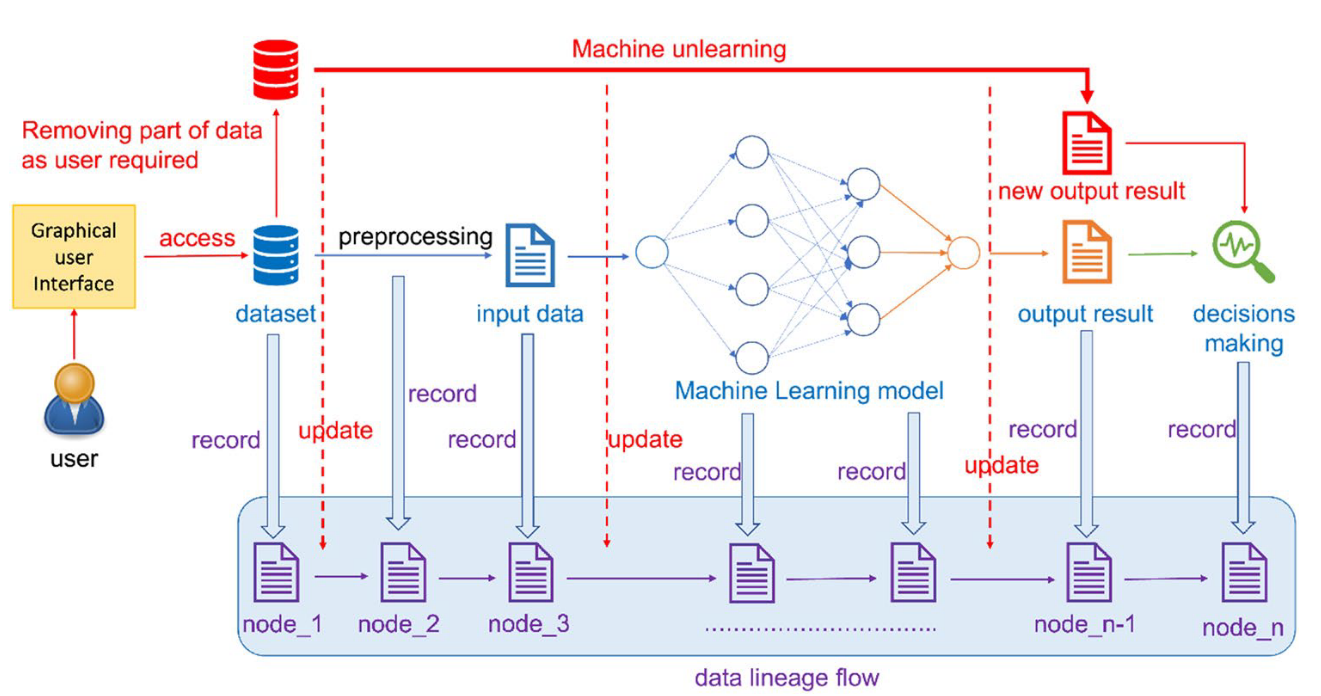}
    \caption{Example of data lineage management for data flow recording and machine unlearning updating (original image is from \cite{Zhang2023})}
    \label{fig:Data lineage}
\end{figure}
How machine unlearning can be applied in conjunction with data lineage management systems is presented in \cite{Zhang2023} as shown in Figure \ref{fig:Data lineage}. Beginning with the user's interaction via a graphical interface, the user can access and potentially request the removal of specific data. This data then undergoes preprocessing before being input into the machine learning model. Throughout each step, from data acquisition to model training and result generation, there are records and updates maintained to capture data flow characteristics and changes. This meticulous recording within the data lineage management system allows developers to have granular control over the model's learning process. When a user requests data removal, the system enters an "unlearning" phase, adjusting the machine learning model and producing a new output result. This comprehensive flow ensures that any alterations, particularly in terms of data removal, are reflected in the decision-making outcomes of the model.
Addressing the biases that may arise during the forgetting process, \cite{Muller2022} developed a forgetfulness stack for data work. This stack operates on records and variables pertinent to data science, aiming to alter human actions throughout the machine learning lifecycle.
\subsubsection{Trustworthiness of forgetting}
%Trustworthiness of forgetting
Incorporating forgetting mechanisms can potentially mitigate social biases in datasets. However, this process introduces human interpretations that may create \textbf{new biases}, termed extrinsic and intrinsic biases. Extrinsic bias emerges from external perceptions of a biased dataset, whereas intrinsic bias arises from modifications within the data itself \cite{Muller2019,Sambasivan2021,Selbst2019,Muller2022}. Forgetting can inadvertently exclude certain groups or perspectives, such as LGBTQIA2S+ individuals or indigenous names, due to system limitations like ASCII character constraints \cite{Costanza-Chock2020,Muller2022}. Moreover, the selective representation in data due to internet access discrepancies further complicates this issue. It's suggested that reintegrating data from underrepresented sources might alleviate these problems \cite{Wegner2021,Vigil-Hayes2017}.

The complexity of machine learning models, especially deep neural networks, makes the \textbf{interpretation of forgetting} mechanisms challenging. These models are often "black boxes" with numerous parameters that obscure how data is forgotten. Adjustments to model architecture, training data modifications, and application of forgetting techniques obscure interpretability further, making it difficult to trace how changes influence the model's predictions \cite{Zhang2023a}.

Originally, machine unlearning aimed to enhance privacy by erasing traces of training data, reducing the risk of privacy leaks. However, recent critiques suggest that machine unlearning could still be vulnerable to attacks such as membership inference attacks, which can potentially reveal sensitive information \cite{Chen2021,Shokri2017,Yeom2018,Sablayrolles2019,Hayes2017}. These challenges underscore the need for robust unlearning methods that can prevent data reconstruction and \textbf{secure privacy} effectively.

\section{Future Directions}
\label{sec7: future directions}
\paragraph{Cross-disciplinary Forgetting}
Cross-disciplinary research is pivotal in advancing machine learning, applying diverse methods to understand forgetting mechanisms and adopting ML techniques to address challenges in other fields. Examples include leveraging insights from neural plasticity and cognitive theories like interference or decay theory \cite{Ebbinghaus2013} to refine ML algorithms. Active collaboration across various fields is essential for innovative breakthroughs.
\paragraph{Forgetting Verification}
Despite various approaches to forgetting in ML, standardised methods for evaluating and verifying their effectiveness are lacking. Developing robust forgetting verification methodologies is critical, especially for scenarios requiring data removal for privacy or compliance reasons. This includes creating algorithms to confirm complete data removal without impacting model performance and ensuring that forgotten data does not indirectly affect future predictions \cite{Nguyen2022}.
\paragraph{Interpretable Forgetting}
The interpretability of forgetting, particularly in neural networks, is challenging. Research could focus on understanding factors affecting transfer learning performance and developing interpretable machine unlearning methods to verify the unlearning process \cite{Basu2020,Koh2017}.
\paragraph{Hybrid Forgetting}
A promising research direction is hybrid forgetting, which combines active and passive forgetting. This approach would allow models to dynamically decide when to update or forget information based on performance, privacy, and compliance, using algorithms that balance data removal's impact on model effectiveness and privacy \cite{Guo2022}.
\paragraph{Source-Free Forgetting}
Source-free forgetting represents a future direction in machine unlearning, moving away from traditional models that require access to all training data for data removal. In stringent privacy settings like the zero-glance setting described by Tarun et al. \cite{Tarun2023}, organisations must erase user data, such as facial recognition details, without reusing those samples even for model adjustments. This necessitates innovative methods to modify model weights and parameters to eliminate data influence without revisiting the data. In transfer learning, similar privacy and regulatory challenges require performing tasks without accessing original source data, a process parallel to zero-glance learning but applied across different domains. Despite progress in reducing negative transfer in source-free transfer learning \cite{Feng2021,Turrisi2022,Liang2020,Ishii2021}, there remains a critical need for algorithms with stronger theoretical underpinnings.
\paragraph{"Goldilocks Zone" of Forgetting}
Identifying the "Goldilocks zone" \footnote{The 'Goldilocks Zone,' or habitable zone, is the area around a star where it is not too hot and not too cold for liquid water to exist on the surface of surrounding planets (NASA). Inspired by \cite{Zhou2022}, we use this term as a metaphor to describe the boundaries and trade-offs involved in forgetting.} of forgetting—finding an optimal balance where a model retains enough information to maintain generalizability while discarding outdated data—is crucial in continual learning. This involves determining the precise threshold for effective learning without catastrophic forgetting.
\paragraph{Trade-offs of forgetting applications}
In industrial applications, implementing forgetting mechanisms involves balancing efficiency, compliance, and historical knowledge retention against challenges like data management complexity and computational costs. Until fully efficient forgetting algorithms are developed, alternative measures like data masking \cite{Archana2018} may be necessary.
\paragraph{Forgetting Regularisation}
Designing forgetting mechanisms involves not only technical challenges but also profound social, political, and ethical considerations \cite{DominguezHernandez2023}. Collaboration with experts across various fields will be crucial to ensure these mechanisms are transparent, fair, and effective, shaping how machine learning impacts our digital society.

\section{Conclusion}
The research outcomes on ``Forgetting'' from other disciplines explain the theoretical foundation and motivation for forgetting in machine learning which tells us \textbf{why to forget}. The process of forgetting in other disciplines, such as cognitive psychology or neuroscience, has been studied extensively, and insights from these fields can be applied to machine learning. For instance, the principles of regularisation, which involve removing or modifying information in machine learning models, are inspired by the concept of forgetting in cognitive psychology. Similarly, the concept of transfer learning, where knowledge from one task is used to improve performance on a different but related task, can be viewed as a form of selective forgetting, where irrelevant information is removed or modified to improve learning efficiency. On the other hand, research from other fields sheds light on \textbf{How to forget} in machine learning. The factors that influence human forgetting behaviour, such as the passage of time, interference, and relevance, can also be implemented in machine learning models. For instance, in machine learning, the decay rate of learned information can be modelled as a function of time, and the degree of interference can be controlled through regularisation techniques. 

Addressing [RQ1], we see that forgetting manifests differently across knowledge areas, each offering unique perspectives that are vital for developing advanced machine learning models. In response to [RQ2], this research classifies and organises the dimensions and approaches of forgetting in ML into a taxonomy to provide clearer literature navigation in this field. Regarding [RQ3], the future of forgetting in machine learning poses both opportunities and challenges, this survey discussed the applications, challenges and concerns and future directions about forgetting in ML.

In conclusion, this survey highlights the critical role of selective forgetting in enhancing machine learning models by enabling them to prioritise relevant information, comply with privacy laws, and mitigate bias. By drawing parallels between human cognitive processes and artificial systems, we open avenues for novel machine learning methodologies that can dynamically adapt and evolve. The exploration of forgetting not only enriches our understanding of machine learning paradigms but also sets the stage for future research that will further refine these concepts, ensuring both ethical and practical advancements in the field.

% Human beings have an advantage over machines when it comes to adaptability and creativity. One of the reasons for this is that the human brain has the ability to forget irrelevant or outdated information, allowing individuals to focus on the most important and relevant aspects of a situation. This forgetting function has been found to facilitate creativity and adaptability in human cognition, allowing individuals to come up with new ideas and solutions based on their experiences. This survey further suggests that machine learning models could benefit from incorporating a similar forgetting function to improve their adaptability and generalisation ability. By selectively forgetting irrelevant or outdated information, machine learning models could become more efficient at identifying and focusing on relevant features of a problem and develop more adaptable solutions based on their learning experiences. 
\pagebreak
\bibliographystyle{ACM-Reference-Format}
\small\bibliography{references}
\end{document}